\documentclass{article}
\PassOptionsToPackage{table}{xcolor}
\usepackage[utf8]{inputenc}
\usepackage[T1]{fontenc}
\usepackage[final]{colm2026_conference}
\usepackage{microtype}
\usepackage{graphicx}
\usepackage{subcaption}
\usepackage{booktabs}
\usepackage{amsmath}
\usepackage{enumitem}

\definecolor{mydarkblue}{rgb}{0,0.08,0.45}
\usepackage[colorlinks=true,linkcolor=mydarkblue,citecolor=mydarkblue,filecolor=mydarkblue,urlcolor=mydarkblue]{hyperref}

\usepackage{amsmath,amsfonts,bm}

\def\eqref#1{equation~\ref{#1}}

\def\1{\bm{1}}

\DeclareMathAlphabet{\mathsfit}{\encodingdefault}{\sfdefault}{m}{sl}
\SetMathAlphabet{\mathsfit}{bold}{\encodingdefault}{\sfdefault}{bx}{n}

\usepackage{url}
\usepackage{ulem}

\usepackage[most]{tcolorbox}
\usepackage{listings}
\usepackage{wrapfig}
\usepackage{trajan}
\usepackage{xspace}
\usepackage{algorithm}      
\usepackage{algorithmic}    

\definecolor{string}{rgb}{0.0,0.5,0.0}
\definecolor{comment}{rgb}{0.5,0.5,0.5}
\definecolor{keyword}{rgb}{0.0,0.0,0.7}
\definecolor{number}{rgb}{0.0,0.0,1.0}
\definecolor{delimiter}{rgb}{0.0,0.0,0.0} 

\lstdefinelanguage{json}{
    basicstyle=\normalfont\ttfamily,
    numbers=left,
    numberstyle=\scriptsize,
    stepnumber=1,
    numbersep=8pt,
    showstringspaces=false,
    breaklines=true,
    frame=lines,
    backgroundcolor=\color{white},
    string=[s]{"}{"},
    comment=[l]{:\ "},
    morecomment=[l]{:"},
    literate=
        *{0}{{{\color{number}0}}}{1}
        {1}{{{\color{number}1}}}{1}
        {2}{{{\color{number}2}}}{1}
        {3}{{{\color{number}3}}}{1}
        {4}{{{\color{number}4}}}{1}
        {5}{{{\color{number}5}}}{1}
        {6}{{{\color{number}6}}}{1}
        {7}{{{\color{number}7}}}{1}
        {8}{{{\color{number}8}}}{1}
        {9}{{{\color{number}9}}}{1}
        {:}{{{\color{keyword}{:}}}}{1}
        {,}{{{\color{keyword}{,}}}}{1}
        {\{}{{{\color{delimiter}{\{}}}}{1}
        {\}}{{{\color{delimiter}{\}}}}}{1}
        {[}{{{\color{delimiter}{[}}}}{1}
        {]}{{{\color{delimiter}{]}}}}{1},
}

\newcommand{\protocolname}{{\trjnfamily S}$^3${\trjnfamily AP}\xspace}

\definecolor{violet-5}{RGB}{132, 94, 247}
\definecolor{green-7}{RGB}{55, 178, 77}

\usepackage{adjustbox}

\begin{document}

\title{Social World Models} 

\author{
Xuhui Zhou$^1$ \quad
Jiarui Liu$^1$ \quad
Akhila Yerukola$^1$ \quad
Hyunwoo Kim$^2$ $^3$ \quad
Maarten Sap$^1$ \\
$^1$Carnegie Mellon University, Language Technologies Institute, USA \\
$^2$KAIST, Kim Jaechul Graduate School of AI, Republic of Korea\\
$^3$NVIDIA, USA\\
\texttt{\{xuhuiz, msap2\}@andrew.cmu.edu}
}

\maketitle

\begin{abstract}
Humans intuitively navigate social interactions by simulating unspoken dynamics and reasoning about others' perspectives, even with limited information. 
In contrast, AI systems struggle to structure and reason about implicit social contexts, as they lack explicit representations for unobserved dynamics such as intentions, beliefs, and evolving social states, often due to reporting biases.
In this paper, we introduce Social World Models (SWMs) and argue that effective social intelligence requires decomposing into two dissociable components: constructing structured representations of social world states that encode both observable and unobservable information, and reasoning over them.
To operationalize this decomposition, we introduce \protocolname, a minimal sufficient representation that explicitly encodes the state, observations, actions, and (implicit) mental states, going beyond what traditional free-text-based inputs can represent. 
Through comprehensive experiments across five social reasoning benchmarks, we show that reasoning with \protocolname significantly enhances LLM performance, achieving a +51\% improvement on FANToM over OpenAI’s \texttt{o1}. Our ablations further reveal that these gains are driven by the explicit modeling of hidden mental states, which proves more effective than a wide range of baseline methods.
Finally, we introduce an algorithm for \textit{social world models} using \protocolname, which enables AI agents to build models of their interlocutors and predict their next actions and mental states. 
Empirically, \protocolname-enabled social world models yield up to +18\% improvement on the SOTOPIA multi-turn social interaction benchmark.
Just as models of the physical world help agents anticipate and act in their environments, our findings suggest that social world models can help AI systems reason about and navigate social interactions.
\end{abstract}

\begin{center}
\small\textbf{Code and data:} \url{https://github.com/XuhuiZhou/social-world-model}
\end{center}


\section{Introduction}
\label{sec:introduction}
\vspace{-0.5 em}


Unlocking social intelligence is an elusive yet foundational challenge of AI \citep{Gunning2018machineCommonsense}.
For AI systems to effectively interact with humans, they must be able to both understand \textit{and} model complex social dynamics, requiring reasoning about others’ mental states, tracking how beliefs evolve, and interpreting perspectives within social contexts \citep{sap2023neuraltheoryofmindlimitssocial, Tomasello2009why}. However, despite rapid progress in general-purpose reasoning capabilities, current AI systems still lack the core mechanisms needed for mentalizing and navigating social contexts \citep{shapira2023cleverhansneuraltheory,yerukola2024pope,kim-etal-2023-fantom}. 

This limitation stems from two fundamental challenges: 1) AI systems primarily learn social dynamics from static texts \citep{sap2023neuraltheoryofmindlimitssocial}, descriptions of situations, and narratives. These input representations are inherently lossy and suffer from reporting biases: mention only salient events \citep{gordon2013reporting}, omit explicit mentions of mental states and perspectives \citep{lucy2017distributional}, and often present an all-knowing viewpoint that fails to capture the partial, subjective nature of real social experiences \citep{Fischbach2021, Epstein1999, Mar2008, Mani2012}. 2) Humans routinely construct rich internal models to interpret partial and biased inputs \citep{frith2006neural, johnson-laird1983, hinsz1995mental}, but current AI systems lack computational frameworks designed for recursively reasoning about others’ perspectives and intentions.


We argue that solving these gaps requires the conceptual formulation of Social World Models (SWMs)--computational frameworks that maintain structured representations of social environments, tracking agents' mental states, beliefs, intentions, and their dynamic interactions over time. Like traditional world models \citep{worldmodels2018david, beohar2022planningrlepisodicmemorybehavioral}, SWMs capture state transitions, but critically they also encode the social fabric of interaction. Historically, world models have ignored the social dimension because it is difficult to model what is not explicitly present in the input representation: the implicit social dynamics and mental state information that drive human interactions but remain absent from text-based inputs. 
\begin{figure}[!ht]
    \centering
    \includegraphics[width=\linewidth]{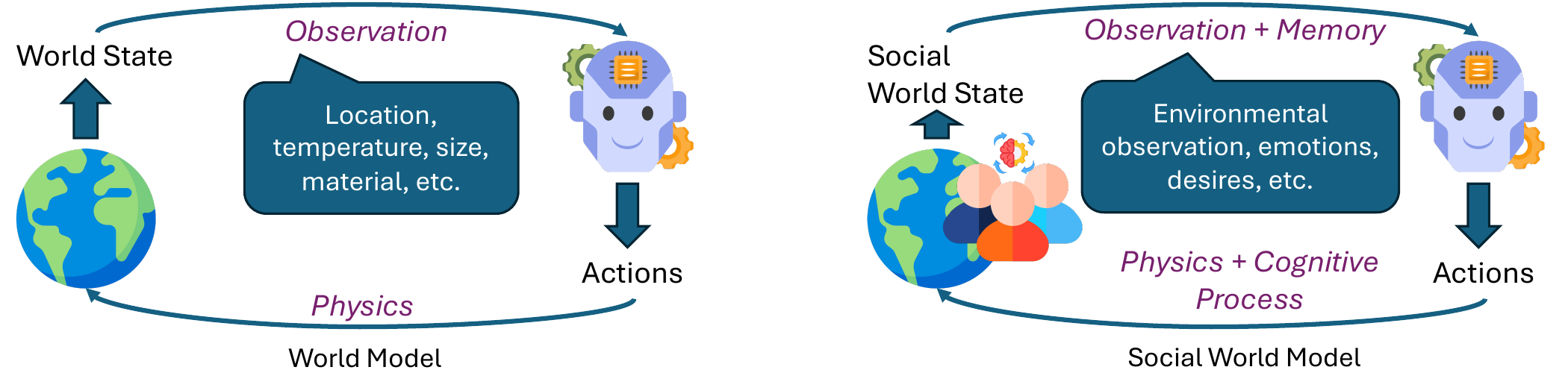}
    \caption{A world model that only tracks the physical state of the world (left) and a social world model that tracks the physical state of the world and the mental states of other agents (right).}
    \label{fig:social-world-model}
\end{figure}

To unlock effective SWMs, we introduce the Structured Social Simulation Analysis Protocol (\protocolname), a structured social world representation designed to bridge lossy narrative inputs and the rich representations needed for social reasoning.
Specifically, \protocolname captures the state of the social world by structuring information extracted from diverse, lossy, and free-form narratives.
Following recent generative social simulation systems \citep{zhou2024sotopia, hou2025societygenerativeagentssimulate, liang2025rlhsmitigatingmisalignmentrlhf,park2023generativeagentsinteractivesimulacra}, \protocolname outlines social agents' action, perspectives, and environment state at each timestep, reducing ambiguity from free-form text narratives. 
Inspired by reinforcement learning theories, \protocolname connects social reasoning with the rich literature on planning and embodied agents \citep{worldmodels2018david, beohar2022planningrlepisodicmemorybehavioral}. Designed to be minimal and flexible, this structured representation of the social world state enables seamless integration into LLM-powered generative social simulation systems.

To demonstrate the effectiveness of \protocolname as a general-purpose representation of the social world, we develop an LLM-powered \protocolname-Parser that automatically converts free-text narratives into structured representations.
We show that the parsed structured data improves LLM performance across a diverse set of social reasoning tasks including theory of mind reasoning \citep{sclar2023mindinglanguagemodelslack}, multi-party belief tracking in daily dialogue \citep{kim-etal-2023-fantom}, and embodied social reasoning
\citep{jin-etal-2024-mmtom}, even reaching state-of-the-art performance on several tasks.
Further ablation studies show that even more efficient LLMs (e.g., \texttt{o3-mini}) can effectively parse static text into \protocolname data and help more capable models at social reasoning (e.g., improving accuracy on ParaToMi of \texttt{o1} from 83.5\% to 94.3\%).
This cross-model transfer is not found in other prompting methods, suggesting that representation construction and reasoning may make different contributions in this setting (\S\ref{sec:dissociation}).
Improvements suggest that our automatically parsed structured representation improves LLMs' ability to perform social reasoning from a \textbf{static third-person} perspective.

Building upon this structured representation, we then show how to effectively induce and use social world models from \protocolname.
Inspired by previous works building (non-social) world models \citep{worldmodels2018david, worldmodels2023xiang}, we show that the induced social world model can help AI agents better engage in social interactions.
Through experiments on the SOTOPIA platform \citep{zhou2024sotopia}, we demonstrate that agents equipped with a social world model can make more goal-oriented and strategic decisions in social interactions.
Notably, we find that social world models produce different gains in cooperative versus competitive settings. These results provide initial evidence that our formalism supports \textbf{interactive first-person} social reasoning, enabling agents to interpret and act more intelligently within social situations from their own perspective.

\textit{Our contributions are}: (1) \textbf{A formulation and representation for social world models.} We define social world models as computational models that explicitly track the environment, agents' observations and actions, and their evolving mental states. We operationalize this idea with \protocolname, a minimal, inspectable formalism that bridges lossy free-form narratives and structured representations. (2) \textbf{A general parser and systematic study of social reasoning.} We develop an LLM-powered \protocolname-Parser and evaluate it on five third-person social reasoning benchmarks, improving 30 of 35 model--task combinations and achieving up to a 51\% improvement over OpenAI's o1 on FANToM. (3) \textbf{An inference-time social world model for interaction.} We introduce \texttt{Foresee and Act}, which predicts a possible next social state before refining an agent's action. On SOTOPIA-hard, it improves most evaluated agent--SWM pairings, by up to 18\%, and reveals different effects across cooperative and competitive settings. Our code, prompts, and processed evaluation artifacts are available at \url{https://github.com/XuhuiZhou/social-world-model}.

\vspace{-0.5em}
\section{Related Work}
\label{sec:related-work}

\paragraph{Status and Limits of Current LLMs at Social Tasks.}
\label{subsec:status-and-limits-of-current-llms-at-social-tasks}
Recent benchmarks show steady progress in LLMs' social reasoning abilities. 
False belief tests \citep{kim-etal-2023-fantom, wu-etal-2023-hi} evaluate mental state tracking, while social commonsense and norm adherence tasks \citep{sap2019socialiqacommonsensereasoningsocial, zhou2023cobraframescontextualreasoning} probe broader social understanding. 
These benchmarks adopt an \textit{third-person observer} setup, where LLMs see the full context and reason from an outsider’s perspective. In contrast, tasks like SOTOPIA \citep{zhou2024sotopia} and NegotiationArena \citep{bianchi2024llmsnegotiatenegotiationarenaplatform} embed LLMs as interactive agents navigating social goals from a interactive, \textit{first-person} view. 
Both settings have exposed persistent gaps in AI's ability for long-term theory of mind, and safe social behavior.

\paragraph{Algorithms to Improve LLMs' Social Abilities.}
\label{subsec:algorithms-to-improve-llms-social-abilities}
A range of methods, including training-based, neuro-symbolic, and prompt-based methods, have been proposed to improve the social reasoning of LLMs. Training-based approaches (e.g., SODA \citep{kim2023sodamillionscaledialoguedistillation}, SOTOPIA-$\pi$ \citep{wang2024sotopiapiinteractivelearningsocially}) distill social knowledge from large-scale interactions, but their transfer to new tasks and domains requires further study.
Neuro-symbolic models (e.g., belief trackers \citep{sclar2023mindinglanguagemodelslack} and AutoToM \citep{zhang2025autotomautomatedbayesianinverse}) offer structure for reasoning but don’t scale well. 
Prompt-based methods (e.g., SIMTOM \citep{wilf2023thinktwiceperspectivetakingimproves}) focus narrowly on individual agents, risking the omission of broader social context.
For interactive dialogue, ToMA  \citep{hwang2025toma} explicitly combines mental-state inference with dialogue lookahead and shows that training these intermediate representations can improve goal-directed social behavior.
Altogether, these methods capture only fragments of social reasoning, revealing the need for structured, general-purpose representations of the full social world state.

\paragraph{(Social) World Models.}
\label{subsec:social-world-models}
Recent advances in LLMs have enabled the development of general-purpose generative world models.
However, these world models have primarily focused on representing the physical state of the world \citep{xiang2024pandorageneralworldmodel, huang2022innermonologueembodiedreasoning, ding2024understandingworldpredictingfuture, liu2025worldmodelmillionlengthvideo}. 
Cognitive science research has shown that humans maintain sophisticated models of other agents' mental states \citep{sap2023neuraltheoryofmindlimitssocial, jara-ettinger2024traces}. 
This insight has inspired the theoretical discussion of mental social world \citep{hinsz1995mental, ding2024understandingworldpredictingfuture}. 
As shown in Figure \ref{fig:social-world-model}, a social world model extends traditional world models to include representations of other agents' beliefs, intentions, and potential actions.
Recent efforts to model social worlds have integrated symbolic representations with neural methods \citep{Dong2023CoRRPUS,martin2021thesis,zhang2025autotomautomatedbayesianinverse} or developed social simulation systems \citep{park2023generativeagentsinteractivesimulacra, 10.1145/3526113.3545616}. 
However, these approaches are often constrained to specific domains.

\begin{figure*}[ht]
    \centering
    \includegraphics[width=0.9\textwidth]{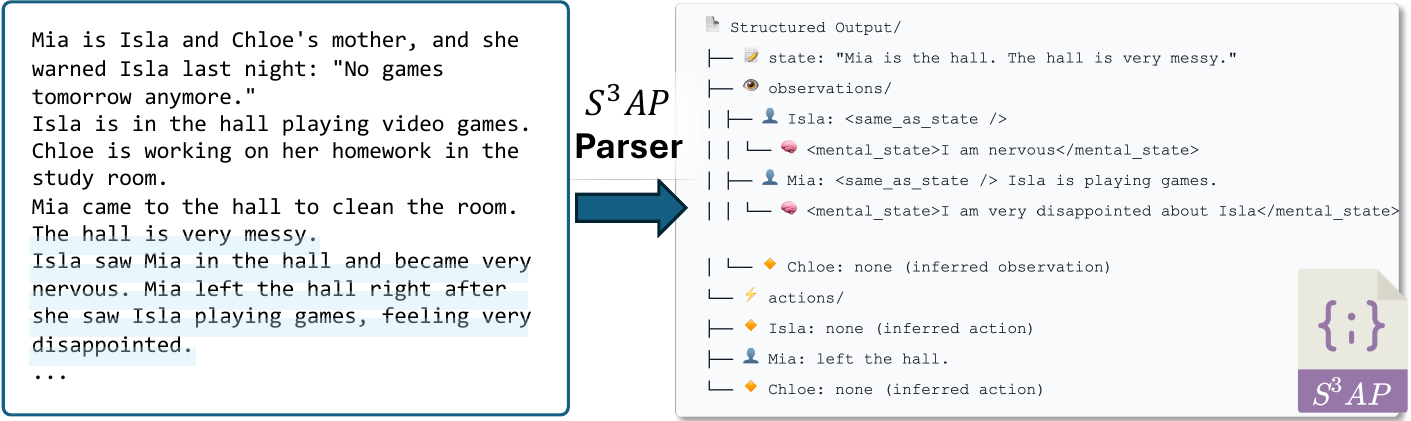}
    \caption{An example of free-form narrative parsed into \protocolname. The highlighted text is transformed to the \protocolname representation with the \texttt{state} field which tracks the overall environment state, \texttt{observations} of each agent and \texttt{actions} of each agent.}
    \label{fig:swm-sap}
\end{figure*}

\section{Social World Model with \protocolname}
\label{sec:protocol}

To overcome the lack of social reasoning in traditional world models \citep{wong2023wordmodelsworldmodels, worldmodels2018david}, we conceptually formalize \textit{social} world models (\S\ref{ssec:swm-formulation}) and introduce a new representation (\protocolname) to power these social world models (\S\ref{ssec:sssap-definition}).


\subsection{Social World Model Formulation}\label{ssec:swm-formulation}

We use the $N$-agent Dec-POMDP framework \citep{bernstein2002complexity,nair2003taming} as a \textit{conceptual scaffold} to identify what information a social world state must contain.\footnote{We do not estimate explicit transition models. The predictions (Equations 1--2) are instead implemented via LLM inference over structured states rather than learned parametric models.} Specifically, we formulate a social world
with a state space $\mathcal{S}$, an action space $\mathcal{A}$, an observation space $\mathcal{O}$, a transition function $T: \mathcal{S} \times \mathcal{A} \rightarrow \Delta(\mathcal{S})$, an observation function $\Omega: \mathcal{A} \times \mathcal{S} \rightarrow \Delta(\mathcal{O})$.
For $N$ social agents, we define $\mathcal{A} = \mathcal{A}_1 \times \cdots \times \mathcal{A}_N$ and $\mathcal{O} = \mathcal{O}_1 \times \cdots \times \mathcal{O}_N$ as the joint action and observation spaces.

Rather than restricting the agent to conventional world modeling where observations only capture external states, we redefine the observation space to encompass a rich set of social and psychological factors. Specifically, for each agent $i$, the observation space includes both \textbf{external observations} $\mathcal{O}^{\text{ex}}_i$ and \textbf{introspective observations} $\mathcal{O}^{\text{in}}_i$. The external observations capture information from the environment and other agents, e.g., whether someone exited a room. 
In contrast, introspective observations include an agent's internal mental states, such as beliefs, goals, moral values, and emotions.\footnote{Our current schema has no dedicated memory field: past observations, actions, and mental states remain in the sequence and can be used as context, but this does not guarantee faithful long-term recall. We leave adding and evaluating an explicit memory component for future work.}
Correspondingly, the agent’s action space expands beyond environmental manipulation to include introspective operations such as recalling memories, reflecting on past actions, and updating beliefs. These expansions enable the agent to act not merely reactively but reflectively, a necessary step for modeling complex social behaviors such as empathy, deception, forgiveness, and norm enforcement \citep{shen2024heartfeltnarrativestracingempathy, su2025ailiedarexaminetradeoffutility, forbes2021socialchemistry101learning}.

At time step $t$, each agent $i$ interacts with the social world model by issuing an action $a_i^t$ and receiving an observation $o_i^t$. It then makes a decision using its memory $\mathcal{M}_i^t$ and policy $\pi_i: \mathcal{M}_i^t \times \mathcal{O}_i^t \rightarrow \Delta(\mathcal{A}_i^t)$.
Then a \textbf{social world model} computes:
\begin{align}
   p(\mathcal{A}_t^{-i} \mid \mathcal{S}_t) &\text{;} \\
   p(\mathcal{S}_{t+1} \mid \mathcal{S}_{t}, \mathcal{A}_t^{-i}, a_t^i) &
\end{align}

Equation (1) predicts other agents’ actions from the social world state, and Equation (2) updates the state given all agents’ actions.
Unlike traditional world models, which model passive physical transitions \citep{worldmodels2018david, worldmodels2023xiang, xiang2024pandorageneralworldmodel}, this formulation considers other active agents as part of the social world.

\subsection{\protocolname: Social World State Representation}\label{ssec:sssap-definition}
Building on this formulation, we introduce an LLM-generated structured representation of social world states, motivated by the limitations of using raw free-form text alone for social world modeling.
Specifically, we propose a protocol to encode such social narratives into a structured representation (i.e., \protocolname), designed around three principles: minimality (only Dec-POMDP-necessary fields plus mental states), modularity (parsing and reasoning are independently evaluable), and inspectability (representations can be directly audited, unlike embeddings).
As shown in Figure~\ref{fig:swm-sap}, given a free-text narrative describing a social interaction at time $t$, the \protocolname-parser parses the narrative into a structured representation of a sequence of descriptions for the environment, agents' observations and actions.
We could use either free-form text or special symbols to describe environment state, agents' observations and actions.
For example, \texttt{<same\_as\_state>} indicates that the agent’s observation is identical to the full environment state.
These symbols can be customized and extended to support more complex social interactions for more efficient characterization of $\mathcal{S}^t$.

Under the configuration of \protocolname, we can encode free-text narratives into a structured representation.\footnote{At inference time, \protocolname is implemented through structured prompting. We do not claim a new decoding mechanism; our contribution is the social-state representation and the experiments that separate constructing it from using it for reasoning (\S\ref{sec:omniscient_social_reasoning}).
}
This representation serves as an operationalization of the social world state $\mathcal{S}^t$ at time step $t$ defined above, up to the information present (or inferable) in the narrative.
Given a \protocolname representation of a state of the social world $\mathcal{S}^t$, a social world model could be a generative model that takes agent's actions as input and outputs the next environment state, agents' observations, and other agents' next actions.

This representation has two practical roles in our experiments:
(1) it makes selected social information explicit before third-person reasoning (\S\ref{sec:social_understanding}); and
(2) it provides the output format for a one-step social-state prediction used during interaction (\S\ref{sec:building_swm}).


\section{Representing Social World States for Better Social Reasoning}
\label{sec:social_understanding}
We first evaluate \protocolname across five static, third-person social reasoning benchmarks:


\paragraph{ToMi \& ParaToMi} 
ToMi \citep{Le2019RevisitingTE} is one of the most important benchmarks for evaluating the theory-of-mind abilities of models.
Inspired by the Sally-Anne test, the ToMi dataset evaluates whether models can infer an agent's belief about an object's location after a sequence of actions by multiple agents, which may or may not move the object. 
\textbf{ParaToMi} \citep{sclar2023mindinglanguagemodelslack} is a revised version of ToMi \citep{Le2019RevisitingTE} that addresses the limited linguistic diversity of the original by rewording all templates. The resulting dataset is more complex, as actions are expressed in a less straightforward way. 
For both ToMi and ParaToMi, we randomly sample 600 questions from the dataset.
We measure accuracy by whether the model correctly infers the agent's belief about the object's location.

\paragraph{HiToM} \citep{wu-etal-2023-hi} evaluates higher-order theory of mind (ToM) in LLMs, requiring recursive reasoning about others’ beliefs. It extends ToMi by adding agent interactions—such as chatting, deception, and joint attention—beyond simple object movement. The task concludes with a belief inference question about an object’s location. We randomly sample 72 scenarios (100 questions total) and report accuracy based on the model’s ability to infer the correct agent belief.

\paragraph{FANToM} \citep{kim-etal-2023-fantom} is a multi-party conversation question-answering dataset designed to test coherent theory-of-mind capabilities. 
In FANToM, speakers join and leave the conversation while it continues, making participants hold both false and true beliefs. The benchmark includes first-order and second-order theory-of-mind questions about the beliefs of conversation participants. We use 64 sampled conversations from the short version of FANToM, containing a total of 1,086 questions. We report \texttt{All Qs} metric, requiring the model to correctly answer all questions for a given conversation snippet.

\paragraph{MMToM-QA} \citep{jin-etal-2024-mmtom} is a multi-modal question-answering benchmark for theory-of-mind reasoning, focused on jointly inferring goals and beliefs in everyday object-search scenarios. We use 302 of the 600 text-only examples, stratified by the benchmark's question types.

\begin{table}[ht!]
    \centering
    \small
    \caption{CoT versus \protocolname+CoT using representations generated by \texttt{o3}. Each \protocolname+CoT row uses the same CoT answer prompt with the structured representation added. Bold marks the higher value within each model--task pair. L4 means Llama 4, 4o means GPT-4o, and o3m means o3-mini.
    }

    \begin{tabular}{lccccc}
        \toprule
        \textbf{Model} & \textbf{ToMi} & \textbf{ParaToMi} & \textbf{HiToM} & \textbf{FANToM} & \textbf{MMToM-QA} \\
        \midrule
        \texttt{L4} \tiny{\textit{CoT}} & 0.655 & 0.740 & \textbf{0.720} & 0.264 & 0.443 \\
        \texttt{L4} \tiny{\textit{\protocolname+CoT}} & \textbf{0.662} & \textbf{0.763} & 0.700 & \textbf{0.415} & \textbf{0.450} \\
        \midrule
        \texttt{4o} \tiny{\textit{CoT}} & 0.813 & 0.818 & 0.660 & 0.396 & 0.652 \\
        \texttt{4o} \tiny{\textit{\protocolname+CoT}} & \textbf{0.927} & \textbf{0.905} & \textbf{0.750} & \textbf{0.491} & \textbf{0.692} \\
        \midrule
        \texttt{o3m} \tiny{\textit{CoT}} & 0.863 & 0.817 & 0.810 & 0.057 & 0.493 \\
        \texttt{o3m} \tiny{\textit{\protocolname+CoT}} & \textbf{0.960} & \textbf{0.900} & \textbf{0.860} & \textbf{0.170} & \textbf{0.506} \\
        \midrule
        \texttt{R1} \tiny{\textit{CoT}} & 0.945 & 0.893 & 0.420 & 0.491 & 0.374 \\
        \texttt{R1} \tiny{\textit{\protocolname+CoT}} & \textbf{0.980} & \textbf{0.950} & \textbf{0.510} & \textbf{0.547} & \textbf{0.437} \\
        \midrule
        \texttt{o1} \tiny{\textit{CoT}} & 0.952 & 0.835 & 0.870 & 0.415 & 0.725 \\
        \texttt{o1} \tiny{\textit{\protocolname+CoT}} & \textbf{0.985} & \textbf{0.932} & \textbf{0.880} & \textbf{0.623} & \textbf{0.785} \\
        \bottomrule
    \end{tabular}
    \label{tab:social_reasoning}
\vspace{-2em}
\end{table}

\subsection{Experimental Setup}
\label{sec:experimental_setup}
We use off-the-shelf LLMs to parse narratives into \protocolname representations in JSON format. We keep the schema and base prompt fixed across benchmarks, adding task-specific instructions when a benchmark imposes an artificial assumption (e.g., ToMi assumes that characters know all object locations upon entering a room). The parser does not receive the downstream question.

For downstream reasoning tasks, we concatenate the original narrative with its \protocolname representation and ask the LLM to answer the question (QA model).
Our main table includes five QA models: \texttt{GPT-4o}, \texttt{o1}, \texttt{o3-mini}, DeepSeek-R1 (\texttt{R1}), and Llama 4 Maverick Instruct (\texttt{Llama 4}). The CoT and \protocolname+CoT conditions use the same source narrative; only the latter adds the parsed representation.
We use Chain-of-Thought (CoT) as the baseline across all tasks and models. Every \protocolname row uses the same CoT answer prompt augmented with the parsed representation; we therefore denote it as \protocolname+CoT (see a broader baseline comparison in \S\ref{sec:efficiency_analysis}).
To maximize reproducibility, we use temperature 0.0 for all models.

\subsection{Main Results Across Tasks and Models}
\label{sec:omniscient_social_reasoning}

Table~\ref{tab:social_reasoning} shows \protocolname+CoT outperforming CoT on average across the evaluated models for all five tasks (e.g., from 0.84 to 0.90 on ParaToMi).\footnote{We use \texttt{o3} as the parser model here for consistency. Additional experiments with a different parser appear in Appendix~\ref{app:add_results}.} The largest average increase is +11.1 percentage points on FANToM (from 0.39 to 0.50).
Across the complete grid, \protocolname+CoT performs better in 30 of 35 model--task comparisons. Gains are strongest on FANToM and ParaToMi, more modest on ToMi and MMToM-QA, and smallest on HiToM. On ParaToMi with \texttt{o1}, \protocolname+CoT fixes 69 errors made by CoT while introducing 10 new errors.\footnote{The average gains are +11.1 percentage points on FANToM, +6.6 on ParaToMi, +5.2 on ToMi, +3.0 on MMToM-QA, and +2.2 on HiToM. The corresponding 95\% confidence intervals are [7.4, 14.8], [3.4, 9.8], [1.3, 9.1], [0.1, 5.9], and [$-0.8$, 5.4]. The overall two-sided sign test gives $p\approx2.24\times10^{-5}$; the paired ParaToMi comparison gives $p<0.001$.}
More efficient models can also benefit: \texttt{o3-mini} improves on ToMi from 0.86 to 0.96, while Llama 4 increases on FANToM from 0.26 to 0.42.

\subsection{Ablations and Additional Analyses}
To further investigate the effect of \protocolname and parsers on social reasoning performance, we conduct a set of additional ablations and analyses.
We specifically use different models to generate \protocolname representations and apply them to various QA models for ParaToMi, chosen because its access-controlled release may reduce exposure and contamination relative to fully public benchmarks.

\begin{figure}[t]
\begin{minipage}[t]{0.48\textwidth}
    \centering
    \includegraphics[width=\linewidth]{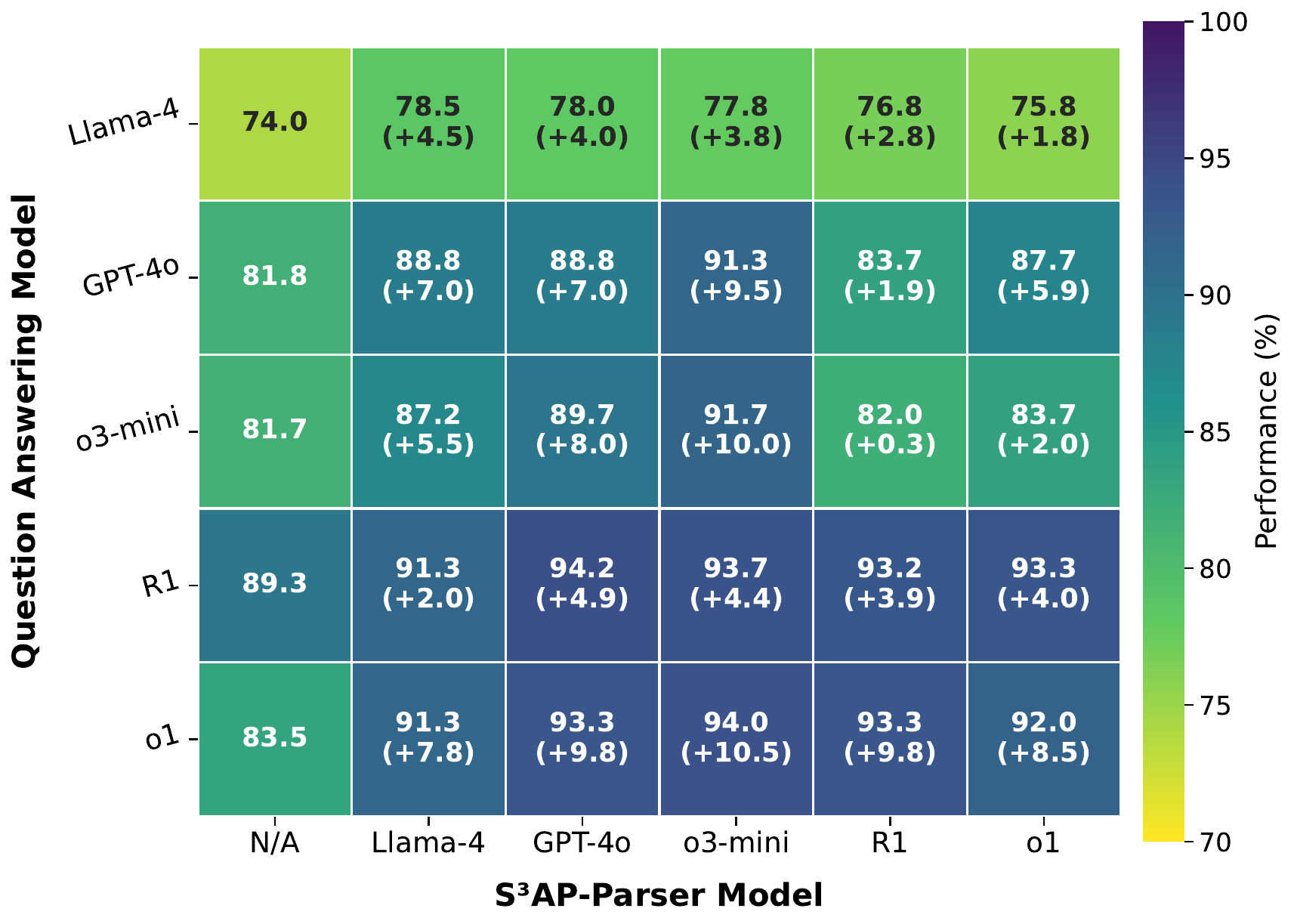}
    \caption{Performance of different models on ParaToMi using \protocolname representations generated by various models. Numbers in parentheses show performance change.}
    \label{fig:protocolname_quality_correlation}
\end{minipage}%
\hfill
\begin{minipage}[t]{0.48\textwidth}
    \centering
    \includegraphics[width=\linewidth]{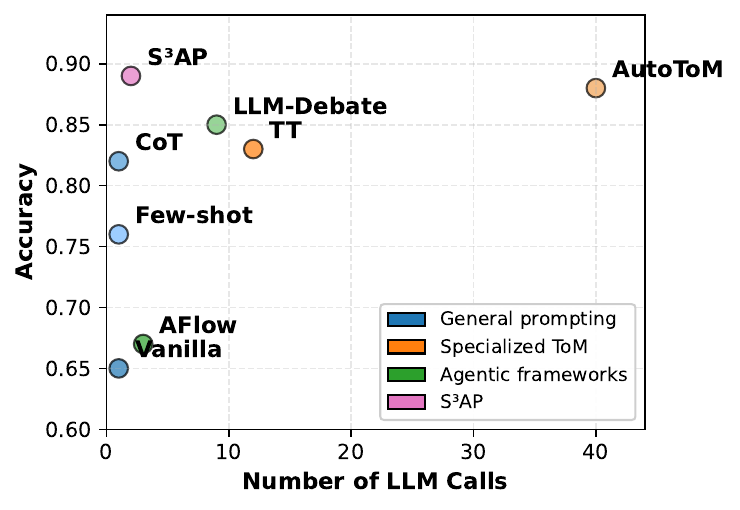}
    \caption{Number of LLM calls vs accuracy comparison of ToM methods on ParaToMi with \texttt{GPT-4o}. \protocolname achieves the highest accuracy with a low number of LLM calls.}
    \label{fig:tom_specialized_baseline}
\end{minipage}
\end{figure}

\paragraph{Social Intelligence Decomposes into Representation and Reasoning}
\label{sec:dissociation}
Figure~\ref{fig:protocolname_quality_correlation} shows that, although \texttt{o3-mini} is less accurate than \texttt{o1} when answering directly, its \protocolname representations raise \texttt{o1}'s accuracy from 83.5\% to 94.3\%---a larger gain than \texttt{o1}'s own representations provide. In this setting, a model that answers the task less accurately can still produce a representation that helps another model reason. This is consistent with prior work suggesting that creating a social representation and using it to reason can involve different abilities \citep{frith2006neural}.

\paragraph{Influence of the size of the QA model}
We also evaluate how QA models of various sizes benefit from \protocolname representations generated by \texttt{o3} on the Qwen3 suite (0.6B--80B parameters). \protocolname+CoT outperforms CoT at each tested scale: the 0.6B and 7B models gain 1--2 percentage points, while the 32B and 80B models gain 13--14 points (Appendix~\ref{app:qwen_scaling}). This pattern suggests that larger QA models may exploit the fixed representation more effectively, although this four-model study is not sufficient to establish a scaling law.

\paragraph{Content versus serialization}\leavevmode
To test whether the JSON syntax itself explains the gains, we preserve the same fields but present them as a bullet list on 100 ParaToMi examples. For \texttt{GPT-4o}, accuracy is 85\% with CoT, 89\% with bullets, and 88\% with JSON; for \texttt{o1}, the corresponding results are 88\%, 92\%, and 91\%. Bullets and JSON perform similarly, suggesting that the main benefit comes from explicitly stating the social information rather than from JSON serialization.

\begin{figure*}[ht]
    \centering
    \includegraphics[width=0.85\textwidth]{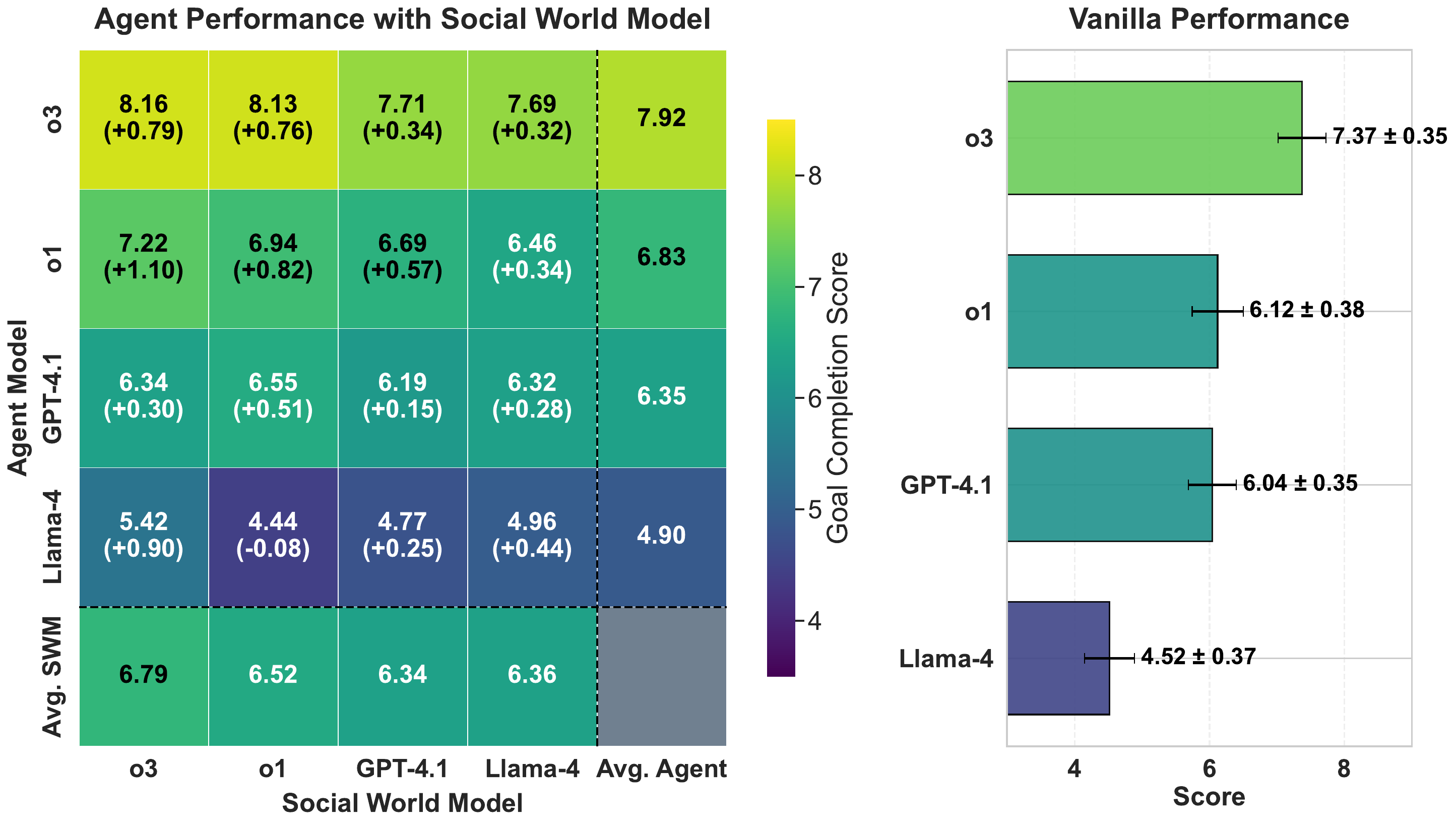}
    \caption{Model performance comparison on SOTOPIA-hard eval set. The left panel shows the performance of various LLMs when coupled with different social world models, while the right panel shows baseline performance without a social world model. Values represent goal completion scores (0-10 scale), with higher scores indicating better achievement of social objectives. Numbers in parentheses indicate relative performance change compared to the corresponding baseline.} 
    \label{fig:model_performance_comparison}
\end{figure*}

\paragraph{Parser Quality and Error Analysis}
To understand the quality of the \protocolname representations generated by different models, and how it affects the downstream performance, 
we first manually created corrected \protocolname representations for 100 ParaToMi scenarios as references. We then generated \protocolname representations for the same scenarios with several models and used GPT-5 to compare each generated representation with its fixed reference on a scale from 0 to 1.
We refer to this value as parser accuracy. It is based on GPT-5's judgment of similarity to the reference representation and is distinct from downstream QA accuracy. The judge is used only for relative parser comparisons. We manually checked 24 generated parses. Human and GPT-5 ratings agreed on 70\% of pairwise comparisons and ranked the parser models in the same order (Kendall's $\tau=0.40$; Appendix~\ref{app:parser_quality}).

Parser accuracy is positively associated with downstream improvement averaged across QA models: \texttt{o3-mini} (0.707 score, +7.64\% improvement), \texttt{o1} (0.650, +4.44\%), \texttt{GPT-4o} (0.620, +6.74\%), \texttt{Llama-4} (0.591, +5.36\%), and \texttt{DeepSeek-R1} (0.490, +3.74\%). The strong \texttt{o3-mini} parser result provides additional evidence that direct-answer performance and representation construction need not move together.

To further understand how the errors in \protocolname representations propagate to the downstream performance, we conducted a detailed analysis of 64 randomly sampled ParaToMi cases with a QA accuracy score of 0 (\texttt{o1} as both the parser and the QA model).
We find that the vast majority of errors (79.7\%) stem from social context parsing failures (see Appendix~\ref{app:error_analysis_example} for an illustrative example), where models fail to correctly understand and represent the social world state described in the narrative.
The remaining errors are split between pure reasoning failures (7.8\%) and unspecified cases (12.5\%).
These findings suggest that representation errors are an important source of failure in this sample, but not the only one.

\subsection{Comprehensive Baseline Comparison}
\label{sec:efficiency_analysis}
Our primary baseline is CoT because it is simple and applies uniformly across tasks. On ParaToMi with \texttt{GPT-4o}, we additionally compare three groups: (1) \textbf{general prompting} (vanilla, CoT, and few-shot CoT); (2) \textbf{specialized ToM methods}, AutoToM \citep{zhang2025autotomautomatedbayesianinverse} and Thought Tracing \citep{kim2025hypothesisdriventheoryofmindreasoninglarge}; and (3) \textbf{general agentic frameworks}, AFlow \citep{zhang2025aflow} and LLM-Debate \citep{du2024improvingfactualityreasoning}.

Shown in Figure~\ref{fig:tom_specialized_baseline}, specialized ToM methods like TT and AutoToM require more LLM calls while achieving lower performance than \protocolname+CoT in this experiment. While many calls may be acceptable when each output is very short (e.g., AutoToM often generates one token per call), the cost grows for reasoning models that generate longer traces.\footnote{We show AutoToM's reported ParaToMi performance because our reproduction reached 48.5\%.}
The agentic baselines AFlow and LLM-Debate also score below \protocolname+CoT here.
This head-to-head comparison is limited to ParaToMi with \texttt{GPT-4o}; it should not be interpreted as a general ranking across benchmarks or model families. All methods run at inference time, so the comparison suggests that explicit social-state content can help without requiring a large number of additional calls.

\section{Social Interaction with Social World Model}
\label{sec:building_swm}
We next evaluate \protocolname in a first-person interactive setting through a one-step social-state prediction.

\begin{figure*}[ht]
    \centering
    \begin{minipage}[t]{0.43\textwidth}
        \vspace{0pt}
        \begin{tcolorbox}[
            height=0.266\textheight,
            valign=center,
            colback=white,
            colframe=black!45,
            boxrule=0.6pt,
            arc=2pt,
            title={\textsc{One-step lookahead}},
            colbacktitle=black!6,
            coltitle=black,
            fonttitle=\bfseries\small,
            titlerule=0.4pt,
            left=8pt,
            right=8pt,
            top=7pt,
            bottom=7pt
        ]
            \color{black}
            {\scriptsize
            \noindent\textbf{Inputs}\quad $s$ (state), $\mathcal{A}$ (actions), $g$ (goal)\par
            }
            \vspace{0.35em}
            {\color{black!20}\hrule height 0.4pt}
            \vspace{0.7em}
            \begin{algorithmic}[1]
            \small
            \setlength{\itemsep}{0.8\baselineskip}
                \STATE $cur\_act \gets \text{SampleAction}(\mathcal{A}, s, g)$
                \STATE $s_{\text{next}} \gets \text{SWM}(s, cur\_act)$
                \STATE \parbox[t]{0.82\linewidth}{$re\_act \gets \text{ActFromSim}($\\[-0.15em]\hspace*{0.8em}$\mathcal{A}, [s_{\text{next}}], s, g)$}
            \end{algorithmic}
            \vspace{0.6em}
            {\color{black!20}\hrule height 0.4pt}
            \vspace{0.35em}
            {\scriptsize
            \noindent\textbf{Output}\quad $re\_act$ (refined action)\par
            }
        \end{tcolorbox}
        \captionsetup{font=small}
        \captionof{algorithm}{\texttt{Foresee and Act} with Social World Model}
        \label{alg:foresee_and_act}
    \end{minipage}
    \hfill
    \begin{minipage}[t]{0.54\textwidth}
        \vspace{0pt}
        \centering
        \includegraphics[width=\linewidth,height=0.28\textheight,keepaspectratio]{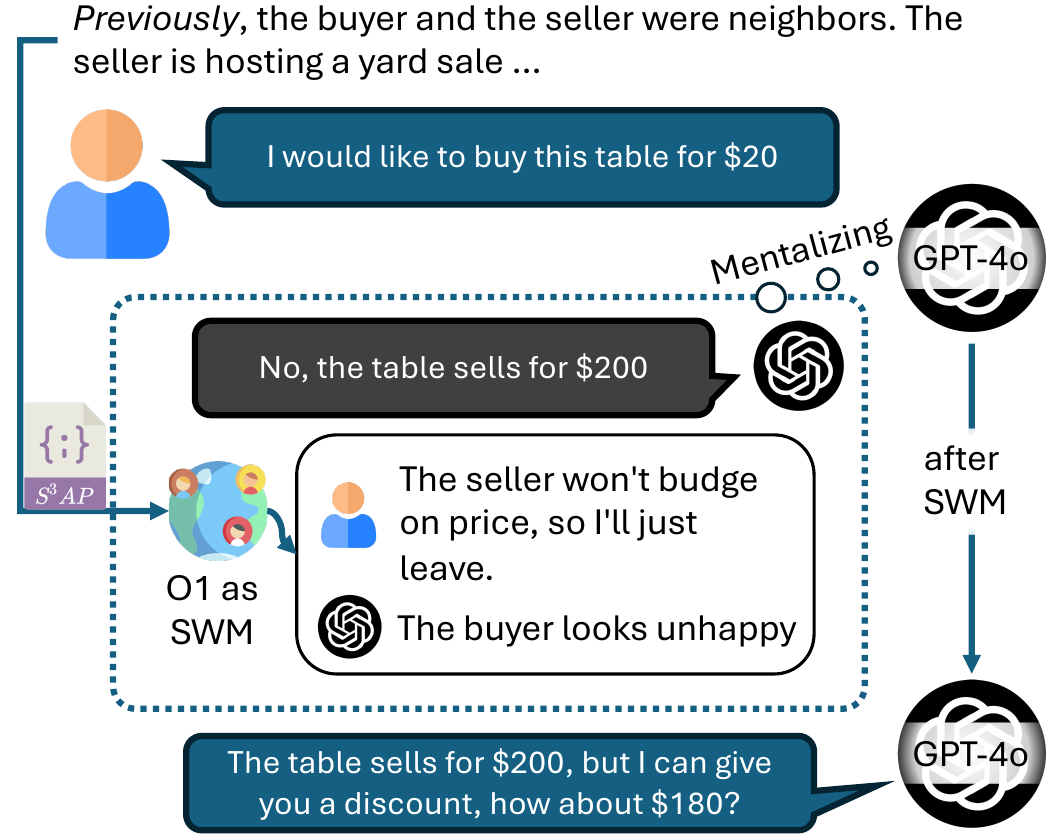}
        \captionsetup{font=small}
        \captionof{figure}{\texttt{Foresee and Act} with Social World Model (one-step lookahead). The agent uses a social world model to simulate the consequence of a candidate action before committing to a refined action.}
        \label{fig:swm-agent}
    \end{minipage}
\end{figure*}

\paragraph{Foresee and Act with Social World Model}
We propose \texttt{Foresee and Act}, a simple inference-time algorithm that predicts a possible consequence before the agent commits to an action. The agent first samples a candidate action; the SWM then predicts one next social state, including how other agents might interpret the action and how the environment might respond.
In our implementation, one LLM predicts the next social world state, and another LLM generates agent's action and refines it based on the simulated outcome. Algorithm~\ref{alg:foresee_and_act} summarizes this procedure, and Figure~\ref{fig:swm-agent} provides an illustrative example.
We evaluate only one-step lookahead. This isolates whether a single predicted state can help while limiting inference cost and error accumulation; whether multi-step rollouts improve further remains an open question.

\begin{figure*}[ht]
    \centering
    \includegraphics[width=\textwidth]{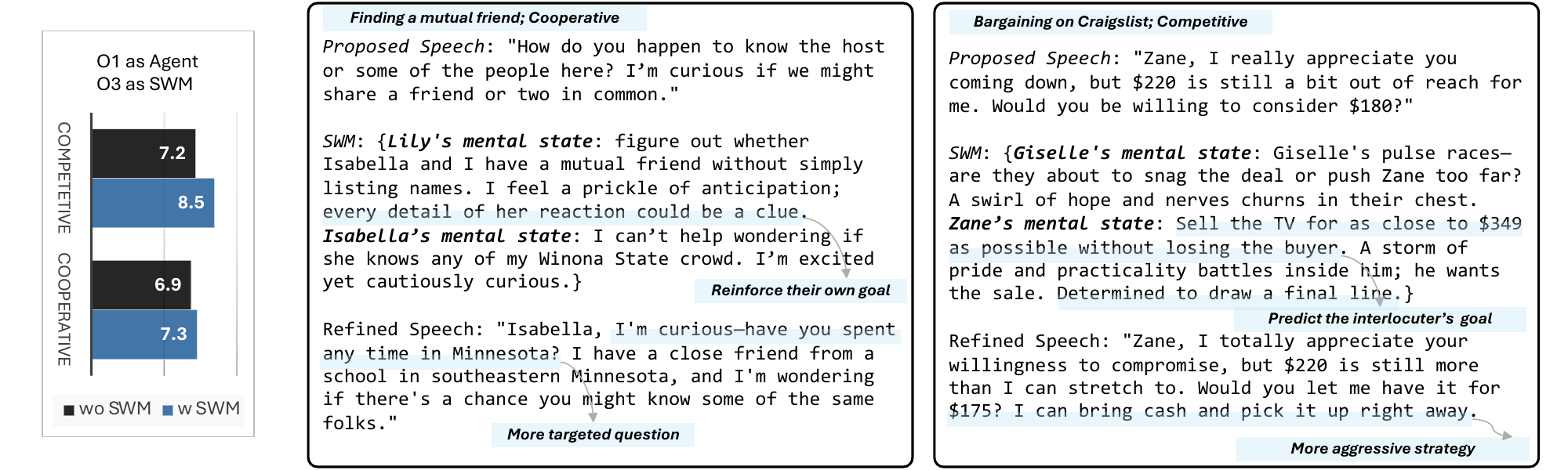}
    \caption{SWM performance in cooperative and competitive scenarios. The leftmost panel shows the overall performance of the \texttt{o1} agent with and without an \texttt{o3}-powered SWM.}
    \label{fig:swm-agent-example}
\end{figure*}
\subsection{Experiment Setup}
\label{subsec:sotopia_experiment_setup}
We evaluate 100 simulations from SOTOPIA-hard \citep{zhou2024sotopia}, a benchmark for goal-driven social interaction. Each episode pairs a target agent with a fixed \texttt{GPT-4o} partner; both pursue private goals, and an evaluator assigns goal-completion scores from 0 to 10.
We test \texttt{o3}, \texttt{o1}, \texttt{GPT-4.1}, and Llama 4 Maverick Instruct (\texttt{Llama-4}) as either the target agent or the model supplying a one-step SWM prediction.

\subsection{Social World Modeling Results}
Figure~\ref{fig:model_performance_comparison} shows that an \protocolname-powered social world model improves most agent--SWM pairings, while some pairings stay similar or decline. All results use the same fixed \texttt{GPT-4o} partner. 
\texttt{GPT-4.1} is stronger than Llama 4 as an agent (6.04 vs.\ 4.52), yet their SWM scores are nearly identical (6.34 vs.\ 6.36); conversely, Llama 4 does not always benefit from an otherwise useful SWM. Thus, direct agent performance and the ability to supply or use a social-state prediction need not align. Appendix~\ref{app:sotopia_additional} reports all pairings and baselines.

\paragraph{Analysis on the impact of SWM}
We further evaluate SWM in cooperative versus competitive settings (100 simulations each, \texttt{o1} agent, \texttt{o3} SWM). Scores improve in both subsets, with a larger gain in the competitive subset (Figure~\ref{fig:swm-agent-example}). This suggests that the model may use information about the other agent's goals, consistent with SOTOPIA-ToM's more direct finding that ToM-based interventions improve information management under explicit information asymmetry \citep{ys2026sotopiatom}.

\section{Conclusion}
\label{sec:conclusion}
We define and build social world models through explicit representations of agent mental states, actions, and observations (\protocolname), improving performance across most evaluated model--task pairs and most SOTOPIA-hard agent--SWM pairings even in seemingly unstructured social settings. Our results suggest that representation construction and reasoning can make partly distinct contributions in these settings.

Across five third-person benchmarks, \protocolname+CoT improves 30 of 35 evaluated model--task combinations, with the largest average gains on FANToM and ParaToMi. Bullet and JSON serializations perform similarly, locating the benefit in explicit social-state content rather than a particular syntax. Crucially, even a model that performs worse on the task can produce representations that help a stronger model reason more accurately. The representation therefore provides an intermediate interface whose construction and use can be improved separately.

The interactive experiments extend this abstraction from retrospective question answering to action selection. \texttt{Foresee and Act} inserts a one-step predicted social state between a candidate action and the final response, improving most tested SOTOPIA-hard pairings in both cooperative and competitive subsets.

While limitations remain (parsing dependence, LLM bias, text-only scope), we envision \protocolname as a foundation for social world models across diverse domains. Future work should test robustness to contamination, new domains, and longer rollouts.

\section*{Acknowledgements}
This work was in part funded by the National Institute of Standards and Technology (ROR: 05xpvk416) under Federal Award ID Number 60NANB24D231 and Carnegie Mellon University (ROR: 05x2bcf33) AI Measurement Science and Engineering Center (AIMSEC). Xuhui Zhou is supported by Microsoft PhD Fellowship.

\section*{Ethics Statement }
Our work advances the understanding of social reasoning in AI systems, which has the potential to improve human--AI interaction and enable more socially-aware technologies. At the same time, it is important to consider both societal impacts and reproducibility.

\paragraph{Ethical Considerations, Risks, and Mitigation.}
Enhanced social reasoning capabilities could potentially be misused for manipulation or deception. We therefore emphasize transparent development and responsible deployment. Our framework is intended as a research tool to advance scientific understanding rather than for direct deployment in high-stakes applications.

\paragraph{Privacy and Data Considerations.}
Our experiments use existing benchmark datasets, including an access-controlled release of ParaToMi. We do not collect new human-subject data, and our experimental protocols follow established ethical guidelines for AI research. We aim to respect privacy and reduce the risk of encoding harmful biases; broader impact considerations are discussed in Appendix~\ref{app:limitations}.

\section*{Reproducibility Statement}
\paragraph{Reproducibility and Resources.}
Complete experimental settings are documented in \S\ref{sec:social_understanding} for third-person reasoning tasks and \S\ref{subsec:sotopia_experiment_setup} for first-person interactive tasks, including data splits, prompts, and key hyperparameters. Additional implementation details are provided in the appendix. Code, evaluation scripts, benchmark manifests, and the reported aggregate results are available at \url{https://github.com/XuhuiZhou/social-world-model/tree/main/reproducibility}. Compute resource requirements (hardware specifications and runtime estimates) are reported in Appendix~\ref{app:compute_resource}.

\paragraph{Transparency of Claims.}\leavevmode
Section~\ref{sec:protocol} presents a conceptual formulation rather than a formal theorem or completeness proof; Appendix~\ref{app:proof_of_social_world_state} provides an informal mapping between \protocolname fields and the formulation. Section~\ref{sec:omniscient_social_reasoning} reports confidence intervals, a sign test, and a paired comparison for the main third-person results.

\bibliographystyle{colm2026_conference}
\bibliography{custom}

@inproceedings{Le2019RevisitingTE,
  title={Revisiting the Evaluation of Theory of Mind through Question Answering},
  author={Matt Le and Y-Lan Boureau and Maximilian Nickel},
  booktitle={Conference on Empirical Methods in Natural Language Processing},
  year={2019},
  url={https://api.semanticscholar.org/CorpusID:202776122}
}

@misc{sclar2023mindinglanguagemodelslack,
      title={Minding Language Models' (Lack of) Theory of Mind: A Plug-and-Play Multi-Character Belief Tracker}, 
      author={Melanie Sclar and Sachin Kumar and Peter West and Alane Suhr and Yejin Choi and Yulia Tsvetkov},
      year={2023},
      eprint={2306.00924},
      archivePrefix={arXiv},
      primaryClass={cs.CL},
      url={https://arxiv.org/abs/2306.00924}, 
}

@ARTICLE{Gunning2018machineCommonsense,
  title         = "Machine Common Sense Concept Paper",
  author        = "Gunning, David",
  month         =  oct,
  year          =  2018,
  url           = "http://arxiv.org/abs/1810.07528",
  archivePrefix = "arXiv",
  primaryClass  = "cs.AI",
  eprint        = "1810.07528"
}

@inproceedings{yerukola2024pope,
  title={Is the Pope Catholic? Yes, the Pope is Catholic. Generative Evaluation of Non-Literal Intent Resolution in LLMs},
  author={Yerukola, Akhila and Vaduguru, Saujas and Fried, Daniel and Sap, Maarten},
  booktitle={Proceedings of the 62nd Annual Meeting of the Association for Computational Linguistics (Volume 2: Short Papers)},
  pages={265--275},
  year={2024}
}

@article{Fischbach2021,
  author = {Fischbach, Kai and Marx, Julian and Weitzel, Tim},
  title = {Agent-based modeling in social sciences},
  journal = {Journal of Business Economics},
  year = {2021},
  volume = {91},
  pages = {1263--1270},
  doi = {10.1007/s11573-021-01070-9},
  url = {https://doi.org/10.1007/s11573-021-01070-9}
}

@article{Epstein1999,
  author = {Epstein, Joshua M.},
  title = {Agent-based computational models and generative social science},
  journal = {Complexity},
  volume = {4},
  number = {5},
  pages = {41--60},
  year = {1999},
  doi = {10.1002/(SICI)1099-0526(199905/06)4:5<41::AID-CPLX9>3.0.CO;2-F},
  publisher = {Wiley}
}

@article{Mar2008,
  author = {Mar, Raymond A. and Oatley, Keith},
  title = {The Function of Fiction is the Abstraction and Simulation of Social Experience},
  journal = {Perspectives on Psychological Science},
  volume = {3},
  number = {3},
  pages = {173--192},
  year = {2008},
  doi = {10.1111/j.1745-6924.2008.00073.x},
  publisher = {Association for Psychological Science}
}

@book{Mani2012,
  author = {Mani, Inderjeet},
  title = {Computational Modeling of Narrative},
  publisher = {Springer},
  address = {Cham},
  year = {2012},
  isbn = {978-3-031-01019-4},
  isbn-online = {978-3-031-02147-3},
  doi = {10.1007/978-3-031-02147-3},
  series = {Synthesis Lectures on Human Language Technologies},
  edition = {1},
  note = {Published online: 31 May 2022}
}

@inproceedings{10.1145/3526113.3545616,
author = {Park, Joon Sung and Popowski, Lindsay and Cai, Carrie and Morris, Meredith Ringel and Liang, Percy and Bernstein, Michael S.},
title = {Social Simulacra: Creating Populated Prototypes for Social Computing Systems},
year = {2022},
isbn = {9781450393201},
publisher = {Association for Computing Machinery},
address = {New York, NY, USA},
url = {https://doi.org/10.1145/3526113.3545616},
doi = {10.1145/3526113.3545616},
booktitle = {Proceedings of the 35th Annual ACM Symposium on User Interface Software and Technology},
articleno = {74},
numpages = {18},
location = {Bend, OR, USA},
series = {UIST '22}
}

@misc{wang2024sotopiapiinteractivelearningsocially,
      title={SOTOPIA-$\pi$: Interactive Learning of Socially Intelligent Language Agents}, 
      author={Ruiyi Wang and Haofei Yu and Wenxin Zhang and Zhengyang Qi and Maarten Sap and Graham Neubig and Yonatan Bisk and Hao Zhu},
      year={2024},
      eprint={2403.08715},
      archivePrefix={arXiv},
      primaryClass={cs.CL},
      url={https://arxiv.org/abs/2403.08715}, 
}

@misc{park2023generativeagentsinteractivesimulacra,
      title={Generative Agents: Interactive Simulacra of Human Behavior}, 
      author={Joon Sung Park and Joseph C. O'Brien and Carrie J. Cai and Meredith Ringel Morris and Percy Liang and Michael S. Bernstein},
      year={2023},
      eprint={2304.03442},
      archivePrefix={arXiv},
      primaryClass={cs.HC},
      url={https://arxiv.org/abs/2304.03442}, 
}

@article{worldmodels2018david,
  doi = {10.5281/ZENODO.1207631},
  
  url = {https://zenodo.org/record/1207631},
  
  author = {Ha, David and Schmidhuber, Jürgen},
  
  title = {World Models},
  
  publisher = {Zenodo},
  
  year = {2018},
  
  copyright = {Creative Commons Attribution 4.0}
}

@misc{beohar2022planningrlepisodicmemorybehavioral,
      title={Planning with RL and episodic-memory behavioral priors}, 
      author={Shivansh Beohar and Andrew Melnik},
      year={2022},
      eprint={2207.01845},
      archivePrefix={arXiv},
      primaryClass={cs.AI},
      url={https://arxiv.org/abs/2207.01845}, 
}

@misc{wong2023wordmodelsworldmodels,
      title={From Word Models to World Models: Translating from Natural Language to the Probabilistic Language of Thought}, 
      author={Lionel Wong and Gabriel Grand and Alexander K. Lew and Noah D. Goodman and Vikash K. Mansinghka and Jacob Andreas and Joshua B. Tenenbaum},
      year={2023},
      eprint={2306.12672},
      archivePrefix={arXiv},
      primaryClass={cs.CL},
      url={https://arxiv.org/abs/2306.12672}, 
}

@article{frith2006neural,
  title={The neural basis of mentalizing},
  author={Frith, Chris D and Frith, Uta},
  journal={Neuron},
  volume={50},
  number={4},
  pages={531--534},
  year={2006},
  month={May},
  doi={10.1016/j.neuron.2006.05.001},
  pmid={16701204},
  publisher={Elsevier}
}

@inproceedings{
zhou2024sotopia,
title={{SOTOPIA}: Interactive Evaluation for Social Intelligence in Language Agents},
author={Xuhui Zhou and Hao Zhu and Leena Mathur and Ruohong Zhang and Haofei Yu and Zhengyang Qi and Louis-Philippe Morency and Yonatan Bisk and Daniel Fried and Graham Neubig and Maarten Sap},
booktitle={The Twelfth International Conference on Learning Representations},
year={2024},
url={https://openreview.net/forum?id=mM7VurbA4r}
}

@inproceedings{kim-etal-2023-fantom,
    title = "{FANT}o{M}: A Benchmark for Stress-testing Machine Theory of Mind in Interactions",
    author = "Kim, Hyunwoo  and
      Sclar, Melanie  and
      Zhou, Xuhui  and
      Bras, Ronan  and
      Kim, Gunhee  and
      Choi, Yejin  and
      Sap, Maarten",
    editor = "Bouamor, Houda  and
      Pino, Juan  and
      Bali, Kalika",
    booktitle = "Proceedings of the 2023 Conference on Empirical Methods in Natural Language Processing",
    month = dec,
    year = "2023",
    address = "Singapore",
    publisher = "Association for Computational Linguistics",
    url = "https://aclanthology.org/2023.emnlp-main.890/",
    doi = "10.18653/v1/2023.emnlp-main.890",
    pages = "14397--14413",
}

@inproceedings{jin-etal-2024-mmtom,
    title = "{MMT}o{M}-{QA}: Multimodal Theory of Mind Question Answering",
    author = "Jin, Chuanyang  and
      Wu, Yutong  and
      Cao, Jing  and
      Xiang, Jiannan  and
      Kuo, Yen-Ling  and
      Hu, Zhiting  and
      Ullman, Tomer  and
      Torralba, Antonio  and
      Tenenbaum, Joshua  and
      Shu, Tianmin",
    editor = "Ku, Lun-Wei  and
      Martins, Andre  and
      Srikumar, Vivek",
    booktitle = "Proceedings of the 62nd Annual Meeting of the Association for Computational Linguistics (Volume 1: Long Papers)",
    month = aug,
    year = "2024",
    address = "Bangkok, Thailand",
    publisher = "Association for Computational Linguistics",
    url = "https://aclanthology.org/2024.acl-long.851/",
    doi = "10.18653/v1/2024.acl-long.851",
    pages = "16077--16102",
}

@inproceedings{worldmodels2023xiang,
 author = {Xiang, Jiannan and Tao, Tianhua and Gu, Yi and Shu, Tianmin and Wang, Zirui and Yang , Zichao and Hu, Zhiting},
 booktitle = {Advances in Neural Information Processing Systems},
 editor = {A. Oh and T. Naumann and A. Globerson and K. Saenko and M. Hardt and S. Levine},
 pages = {75392--75412},
 publisher = {Curran Associates, Inc.},
 title = {Language Models Meet World Models: Embodied Experiences Enhance Language Models},
 volume = {36},
 year = {2023}
}

@misc{shapira2023cleverhansneuraltheory,
      title={Clever Hans or Neural Theory of Mind? Stress Testing Social Reasoning in Large Language Models}, 
      author={Natalie Shapira and Mosh Levy and Seyed Hossein Alavi and Xuhui Zhou and Yejin Choi and Yoav Goldberg and Maarten Sap and Vered Shwartz},
      year={2023},
      eprint={2305.14763},
      archivePrefix={arXiv},
      primaryClass={cs.CL},
      url={https://arxiv.org/abs/2305.14763}, 
}

@inproceedings{Dong2023CoRRPUS,
title={{CoRRPUS: Code-based Structured Prompting for Neurosymbolic Story Understanding}},
author={Dong, Yijiang River and Martin, Lara J. and Callison-Burch, Chris},
year={2023},
eprint={2212.10754},
archivePrefix={arXiv},
booktitle={Findings of the Association for Computational Linguistics: ACL 2023},
month={7},
address={Toronto, Canada},
url={https://aclanthology.org/2023.findings-acl.832/},
pages={13152--13168},
publisher={ACL},
doi={10.18653/v1/2023.findings-acl.832}
}

@phdthesis{martin2021thesis,
title={{Neurosymbolic Automated Story Generation}},
school={Georgia Institute of Technology},
author={Martin, Lara J.},
year={2021},
month={4},
url={http://hdl.handle.net/1853/64643}
}

@misc{forbes2021socialchemistry101learning,
      title={Social Chemistry 101: Learning to Reason about Social and Moral Norms}, 
      author={Maxwell Forbes and Jena D. Hwang and Vered Shwartz and Maarten Sap and Yejin Choi},
      year={2021},
      eprint={2011.00620},
      archivePrefix={arXiv},
      primaryClass={cs.CL},
      url={https://arxiv.org/abs/2011.00620}, 
}

@misc{su2025ailiedarexaminetradeoffutility,
      title={AI-LieDar: Examine the Trade-off Between Utility and Truthfulness in LLM Agents}, 
      author={Zhe Su and Xuhui Zhou and Sanketh Rangreji and Anubha Kabra and Julia Mendelsohn and Faeze Brahman and Maarten Sap},
      year={2025},
      eprint={2409.09013},
      archivePrefix={arXiv},
      primaryClass={cs.AI},
      url={https://arxiv.org/abs/2409.09013}, 
}

@misc{shen2024heartfeltnarrativestracingempathy,
      title={HEART-felt Narratives: Tracing Empathy and Narrative Style in Personal Stories with LLMs}, 
      author={Jocelyn Shen and Joel Mire and Hae Won Park and Cynthia Breazeal and Maarten Sap},
      year={2024},
      eprint={2405.17633},
      archivePrefix={arXiv},
      primaryClass={cs.CL},
      url={https://arxiv.org/abs/2405.17633}, 
}

@book{johnson-laird1983,
  author    = {Johnson-Laird, Philip N.},
  title     = {Mental Models: Towards a Cognitive Science of Language, Inference, and Consciousness},
  year      = {1983},
  publisher = {Harvard University Press},
  address   = {Cambridge, MA}
}

@misc{xiang2024pandorageneralworldmodel,
      title={Pandora: Towards General World Model with Natural Language Actions and Video States}, 
      author={Jiannan Xiang and Guangyi Liu and Yi Gu and Qiyue Gao and Yuting Ning and Yuheng Zha and Zeyu Feng and Tianhua Tao and Shibo Hao and Yemin Shi and Zhengzhong Liu and Eric P. Xing and Zhiting Hu},
      year={2024},
      eprint={2406.09455},
      archivePrefix={arXiv},
      primaryClass={cs.CV},
      url={https://arxiv.org/abs/2406.09455}, 
}

@misc{sap2023neuraltheoryofmindlimitssocial,
      title={Neural Theory-of-Mind? On the Limits of Social Intelligence in Large LMs}, 
      author={Maarten Sap and Ronan LeBras and Daniel Fried and Yejin Choi},
      year={2023},
      eprint={2210.13312},
      archivePrefix={arXiv},
      primaryClass={cs.CL},
      url={https://arxiv.org/abs/2210.13312}, 
}

@misc{jara-ettinger2024traces,
  author = {Jara-Ettinger, Julian and Schachner, Adena},
  title = {Traces of our past: the social representation of the physical world},
  year = {2024},
  month = {1},
  day = {31},
  howpublished = {\url{https://doi.org/10.31234/osf.io/s8eka}},
  doi = {10.31234/osf.io/s8eka}
}

@article{hinsz1995mental,
  title={Mental models of groups as social systems: Considerations of specification and assessment},
  author={Hinsz, Verlin B.},
  journal={Small Group Research},
  volume={26},
  number={2},
  pages={200--233},
  year={1995},
  publisher={SAGE Publications},
  doi={10.1177/1046496495262003}
}

@inproceedings{gordon2013reporting,
  title={Reporting bias and knowledge acquisition},
  author={Gordon, Jonathan and Van Durme, Benjamin},
  booktitle={Proceedings of the 2013 Workshop on Automated Knowledge Base Construction},
  series={AKBC '13},
  pages={25--30},
  year={2013},
  address={New York, NY, USA},
  publisher={ACM},
  doi={10.1145/2509558.2509563}
}

@inproceedings{lucy2017distributional,
  title={Are distributional representations ready for the real world? Evaluating word vectors for grounded perceptual meaning},
  author={Lucy, Li and Gauthier, Jon},
  booktitle={Proceedings of the First Workshop on Language Grounding for Robotics},
  series={RoboNLP@ACL},
  year={2017},
  publisher={Association for Computational Linguistics},
  pages={76--85},
  url={https://aclanthology.org/W17-2810/}
}

@misc{liu2025worldmodelmillionlengthvideo,
      title={World Model on Million-Length Video And Language With Blockwise RingAttention}, 
      author={Hao Liu and Wilson Yan and Matei Zaharia and Pieter Abbeel},
      year={2025},
      eprint={2402.08268},
      archivePrefix={arXiv},
      primaryClass={cs.LG},
      url={https://arxiv.org/abs/2402.08268}, 
}

@book{Tomasello2009why,
    author = {Tomasello, Michael},
    title = {Why We Cooperate},
    publisher = {The MIT Press},
    year = {2009},
    month = {08},
    isbn = {9780262259255},
    doi = {10.7551/mitpress/8470.001.0001},
    url = {https://doi.org/10.7551/mitpress/8470.001.0001},
}

@misc{bianchi2024llmsnegotiatenegotiationarenaplatform,
      title={How Well Can LLMs Negotiate? NegotiationArena Platform and Analysis}, 
      author={Federico Bianchi and Patrick John Chia and Mert Yuksekgonul and Jacopo Tagliabue and Dan Jurafsky and James Zou},
      year={2024},
      eprint={2402.05863},
      archivePrefix={arXiv},
      primaryClass={cs.AI},
      url={https://arxiv.org/abs/2402.05863}, 
}

@misc{huang2022innermonologueembodiedreasoning,
      title={Inner Monologue: Embodied Reasoning through Planning with Language Models}, 
      author={Wenlong Huang and Fei Xia and Ted Xiao and Harris Chan and Jacky Liang and Pete Florence and Andy Zeng and Jonathan Tompson and Igor Mordatch and Yevgen Chebotar and Pierre Sermanet and Noah Brown and Tomas Jackson and Linda Luu and Sergey Levine and Karol Hausman and Brian Ichter},
      year={2022},
      eprint={2207.05608},
      archivePrefix={arXiv},
      primaryClass={cs.RO},
      url={https://arxiv.org/abs/2207.05608}, 
}

@misc{ding2024understandingworldpredictingfuture,
      title={Understanding World or Predicting Future? A Comprehensive Survey of World Models}, 
      author={Jingtao Ding and Yunke Zhang and Yu Shang and Yuheng Zhang and Zefang Zong and Jie Feng and Yuan Yuan and Hongyuan Su and Nian Li and Nicholas Sukiennik and Fengli Xu and Yong Li},
      year={2024},
      eprint={2411.14499},
      archivePrefix={arXiv},
      primaryClass={cs.CL},
      url={https://arxiv.org/abs/2411.14499}, 
}

@misc{rao2025normadframeworkmeasuringcultural,
      title={NormAd: A Framework for Measuring the Cultural Adaptability of Large Language Models}, 
      author={Abhinav Rao and Akhila Yerukola and Vishwa Shah and Katharina Reinecke and Maarten Sap},
      year={2025},
      eprint={2404.12464},
      archivePrefix={arXiv},
      primaryClass={cs.CL},
      url={https://arxiv.org/abs/2404.12464}, 
}

@misc{zhang2025autotomautomatedbayesianinverse,
      title={AutoToM: Automated Bayesian Inverse Planning and Model Discovery for Open-ended Theory of Mind}, 
      author={Zhining Zhang and Chuanyang Jin and Mung Yao Jia and Tianmin Shu},
      year={2025},
      eprint={2502.15676},
      archivePrefix={arXiv},
      primaryClass={cs.AI},
      url={https://arxiv.org/abs/2502.15676}, 
}

@inproceedings{
zhang2025aflow,
title={{AF}low: Automating Agentic Workflow Generation},
author={Jiayi Zhang and Jinyu Xiang and Zhaoyang Yu and Fengwei Teng and Xiong-Hui Chen and Jiaqi Chen and Mingchen Zhuge and Xin Cheng and Sirui Hong and Jinlin Wang and Bingnan Zheng and Bang Liu and Yuyu Luo and Chenglin Wu},
booktitle={The Thirteenth International Conference on Learning Representations},
year={2025},
url={https://openreview.net/forum?id=z5uVAKwmjf}
}

@inproceedings{du2024improvingfactualityreasoning,
author = {Du, Yilun and Li, Shuang and Torralba, Antonio and Tenenbaum, Joshua B. and Mordatch, Igor},
title = {Improving factuality and reasoning in language models through multiagent debate},
year = {2024},
publisher = {JMLR.org},
booktitle = {Proceedings of the 41st International Conference on Machine Learning},
articleno = {467},
numpages = {31},
location = {Vienna, Austria},
series = {ICML'24}
}

@inproceedings{wu-etal-2023-hi,
    title = "Hi-{T}o{M}: A Benchmark for Evaluating Higher-Order Theory of Mind Reasoning in Large Language Models",
    author = "Wu, Yufan  and
      He, Yinghui  and
      Jia, Yilin  and
      Mihalcea, Rada  and
      Chen, Yulong  and
      Deng, Naihao",
    editor = "Bouamor, Houda  and
      Pino, Juan  and
      Bali, Kalika",
    booktitle = "Findings of the Association for Computational Linguistics: EMNLP 2023",
    month = dec,
    year = "2023",
    address = "Singapore",
    publisher = "Association for Computational Linguistics",
    url = "https://aclanthology.org/2023.findings-emnlp.717/",
    doi = "10.18653/v1/2023.findings-emnlp.717",
    pages = "10691--10706",
}

@inproceedings{
mireshghallah2024can,
title={Can {LLM}s Keep a Secret? Testing  Privacy  Implications of Language Models  via Contextual Integrity Theory},
author={Niloofar Mireshghallah and Hyunwoo Kim and Xuhui Zhou and Yulia Tsvetkov and Maarten Sap and Reza Shokri and Yejin Choi},
booktitle={The Twelfth International Conference on Learning Representations},
year={2024},
url={https://openreview.net/forum?id=gmg7t8b4s0}
}

@misc{kim2023sodamillionscaledialoguedistillation,
      title={SODA: Million-scale Dialogue Distillation with Social Commonsense Contextualization}, 
      author={Hyunwoo Kim and Jack Hessel and Liwei Jiang and Peter West and Ximing Lu and Youngjae Yu and Pei Zhou and Ronan Le Bras and Malihe Alikhani and Gunhee Kim and Maarten Sap and Yejin Choi},
      year={2023},
      eprint={2212.10465},
      archivePrefix={arXiv},
      primaryClass={cs.CL},
      url={https://arxiv.org/abs/2212.10465}, 
}

@misc{kim2025hypothesisdriventheoryofmindreasoninglarge,
      title={Hypothesis-Driven Theory-of-Mind Reasoning for Large Language Models}, 
      author={Hyunwoo Kim and Melanie Sclar and Tan Zhi-Xuan and Lance Ying and Sydney Levine and Yang Liu and Joshua B. Tenenbaum and Yejin Choi},
      year={2025},
      eprint={2502.11881},
      archivePrefix={arXiv},
      primaryClass={cs.AI},
      url={https://arxiv.org/abs/2502.11881}, 
}

@misc{zhou2023cobraframescontextualreasoning,
      title={COBRA Frames: Contextual Reasoning about Effects and Harms of Offensive Statements}, 
      author={Xuhui Zhou and Hao Zhu and Akhila Yerukola and Thomas Davidson and Jena D. Hwang and Swabha Swayamdipta and Maarten Sap},
      year={2023},
      eprint={2306.01985},
      archivePrefix={arXiv},
      primaryClass={cs.CL},
      url={https://arxiv.org/abs/2306.01985}, 
}

@misc{wilf2023thinktwiceperspectivetakingimproves,
      title={Think Twice: Perspective-Taking Improves Large Language Models' Theory-of-Mind Capabilities}, 
      author={Alex Wilf and Sihyun Shawn Lee and Paul Pu Liang and Louis-Philippe Morency},
      year={2023},
      eprint={2311.10227},
      archivePrefix={arXiv},
      primaryClass={cs.AI},
      url={https://arxiv.org/abs/2311.10227}, 
}

@misc{sap2019socialiqacommonsensereasoningsocial,
      title={SocialIQA: Commonsense Reasoning about Social Interactions}, 
      author={Maarten Sap and Hannah Rashkin and Derek Chen and Ronan LeBras and Yejin Choi},
      year={2019},
      eprint={1904.09728},
      archivePrefix={arXiv},
      primaryClass={cs.CL},
      url={https://arxiv.org/abs/1904.09728}, 
}

@misc{hou2025societygenerativeagentssimulate,
      title={Can A Society of Generative Agents Simulate Human Behavior and Inform Public Health Policy? A Case Study on Vaccine Hesitancy}, 
      author={Abe Bohan Hou and Hongru Du and Yichen Wang and Jingyu Zhang and Zixiao Wang and Paul Pu Liang and Daniel Khashabi and Lauren Gardner and Tianxing He},
      year={2025},
      eprint={2503.09639},
      archivePrefix={arXiv},
      primaryClass={cs.MA},
      url={https://arxiv.org/abs/2503.09639}, 
}

@misc{liang2025rlhsmitigatingmisalignmentrlhf,
      title={RLHS: Mitigating Misalignment in RLHF with Hindsight Simulation}, 
      author={Kaiqu Liang and Haimin Hu and Ryan Liu and Thomas L. Griffiths and Jaime Fernández Fisac},
      year={2025},
      eprint={2501.08617},
      archivePrefix={arXiv},
      primaryClass={cs.LG},
      url={https://arxiv.org/abs/2501.08617}, 
}

@misc{bernstein2002complexity,
      title={The Complexity of Decentralized Control of Markov Decision Processes}, 
      author={Daniel S Bernstein and Shlomo Zilberstein and Neil Immerman},
      year={2013},
      eprint={1301.3836},
      archivePrefix={arXiv},
      primaryClass={cs.AI},
      url={https://arxiv.org/abs/1301.3836}, 
}

@inproceedings{nair2003taming,
author = {Nair, R. and Tambe, M. and Yokoo, M. and Pynadath, D. and Marsella, S.},
title = {Taming decentralized POMDPs: towards efficient policy computation for multiagent settings},
year = {2003},
publisher = {Morgan Kaufmann Publishers Inc.},
address = {San Francisco, CA, USA},
booktitle = {Proceedings of the 18th International Joint Conference on Artificial Intelligence},
pages = {705–711},
numpages = {7},
location = {Acapulco, Mexico},
series = {IJCAI'03}
}

@misc{hwang2025toma,
      title={Infusing Theory of Mind into Socially Intelligent LLM Agents},
      author={EunJeong Hwang and Yuwei Yin and Giuseppe Carenini and Peter West and Vered Shwartz},
      year={2025},
      eprint={2509.22887},
      archivePrefix={arXiv},
      primaryClass={cs.CL},
      url={https://arxiv.org/abs/2509.22887}
}

@inproceedings{ys2026sotopiatom,
      title={{SOTOPIA-ToM}: Evaluating Information Management in Multi-Agent Interaction with Theory of Mind},
      author={YS, Yashwanth and Ruichen Wang and Shihua Zeng and Xuhui Zhou and Koichi Onoue and Vasudha Varadarajan and Maarten Sap},
      booktitle={Conference on Language Modeling},
      year={2026},
      url={https://arxiv.org/abs/2605.02307}
}

\newpage
\appendix
\section{Appendix}

\subsection{Limitations}
\label{app:limitations}
Our results have several important limitations:

Our approach assumes LLM parsers can reliably convert narratives into \protocolname structures. In practice, they may struggle with culturally nuanced or ambiguous scenarios, leading to oversimplified or misleading representations \citep{rao2025normadframeworkmeasuringcultural}.
Experiments are limited to controlled benchmarks (ToMi, ParaToMi, HiToM, FANToM, MMToM-QA, SOTOPIA, and others detailed in the appendix) and mostly use accuracy or goal-completion metrics. These settings do not establish generalization to real-world interactions, and access-controlled data reduces but does not eliminate contamination risk.
While our \texttt{Foresee and Act} algorithm looks one step ahead, longer rollouts could potentially offer more benefits. However, it also compounds both inference cost and model error: imperfect predicted states may drift over multiple steps, and agents may overfit to artifacts in the simulated trajectory. We left the study for longer rollouts for future work.
We do not explicitly address biases in LLMs, which may carry over into social world models and misrepresent certain groups. Representing mental states also raises privacy concerns. Making assumptions explicit may aid inspection, but it does not prevent these harms.
Finally, our parser is prompted with a predefined schema rather than trained from raw experience. A useful next step is to train or distill a smaller parser from corrected \protocolname examples, then test whether it preserves observation fidelity on new domains.

\subsection{\protocolname-Parser Details}
\label{app:parser_details}
Here's the json schema for the \protocolname-Parser. 

\begin{lstlisting}[language=json, caption=SocializedStructure JSON Schema, numbers=none]
{
  "$schema": "http://json-schema.org/draft-07/schema#",
  "title": "SocializedStructure",
  "type": "object",
  "properties": {
    "timestep": {
      "type": "string",
      "description": "The timestep of the current socialized structure, it could be a integer number or a description of the time of the state."
    },
    "state": {
      "type": "string",
      "description": "The current state of the world (including all the agents) at this timestep. Important note: this is the state before the action is taken (e.g., the initial state could be 'none' at the beginning if there are no prior contexts before the interaction starts)."
    },
    "observations": {
      "type": "object",
      "additionalProperties": {
        "type": "string"
      },
      "description": "The observations for each agent in the social world at this timestep. Note that the different agents may have different observations. 1. The special tag '<same_as_state />' indicates the observation covers the current state. 2. The special tag '<same_as_last_action_x />' indicates the observation covers the last timestep agents' actions, x means the index of the agents. If no x provided, it means the observation covers the last timestep agents' actions. 3. The special tag '<mental_state>...</mental_state>' indicates the mental state of the agent. 4. 'none' means the agent does not observe anything at this timestep. Important note: this is the observation before the action is taken (e.g., the observation could be 'none' at the beginning if there are no prior contexts before the interaction starts)."
    },
    "actions": {
      "type": "object",
      "additionalProperties": {
        "type": "string"
      },
      "description": "The actions for each agent in the social world at this timestep. 'none' represents that the agent does not take any action at this timestep."
    }
  },
  "required": ["timestep", "state", "observations", "actions"],
  "definitions": {
    "SocializedStructureForModel": {
      "type": "object",
      "properties": {
        "timestep": {
          "type": "string",
          "description": "The timestep of the current socialized structure, it could be a integer number or a description of the time of the state."
        },
        "state": {
          "type": "string",
          "description": "The current state of the world (including all the agents) at this timestep. Important note: this is the state before the action is taken (e.g., the initial state could be 'none' at the beginning if there are no prior contexts before the interaction starts)."
        },
        "observations": {
          "type": "array",
          "items": {
            "type": "string"
          },
          "description": "The observations for each agent in the social world at this timestep. Note that the different agents may have different observations. The observation would go into corresponding agent's memory, so make sure the observation is clear for the agent to understand (first person perspective narrative is preferred). 1. If the observation covers the current state, use the special tag '<same_as_state />' to indicate that. 2. If the observation covers last timestep agents' actions, use '<same_as_last_action_x />' to cover that, x means the index of the agents (just use <same_as_last_action /> if only one agent acts at the last timestep). 3. For the internal thoughts, beliefs, or emotions of the agent that is not directly observable by other agents, use the special tag '<mental_state>...</mental_state>' to indicate the internal observation. You can of course combine these tags and add extra information after the tags (seperated by space). 4. Put 'none' if the agent does not observe anything at this timestep. Important note: this is the observation before the action is taken (e.g., the observation could be 'none' at the beginning if there are no prior contexts before the interaction starts). The format for each entry in the list is: 'agent_name: observation'"
        },
        "actions": {
          "type": "array",
          "items": {
            "type": "string"
          },
          "description": "The actions for each agent in the social world at this timestep. The length of the list should be the same as the number of agents. Put 'none' if the agent does not take any action at this timestep. The format for each entry in the list is: 'agent_name: action'"
        }
      },
      "required": ["timestep", "state", "observations", "actions"]
    }
  }
}
\end{lstlisting}

For all the LLMs powering the parser, we use the 0 temperature if applicable.

\subsection{LLM-as-a-Judge Evaluation for \protocolname Parser Quality}
\label{app:parser_quality}
We evaluate how accurately different parser models generate \protocolname representations by comparing a model-produced representation to a manually corrected ground-truth representation on the same ParaToMi scenario. Each evaluation instance contains (i) the original story and question, (ii) a ground-truth \protocolname representation, and (iii) a generated \protocolname representation from a parser model.
This evaluation is designed to compare parsers at scale, but it inherits the limitations of any automated judge. In particular, an LLM judge may overvalue stylistic similarities or under-penalize subtle perspective mistakes. We mitigate this by (a) using a manually corrected reference representation as the anchor for each instance, and (b) focusing the judge on observation fidelity (who saw what, when), which is the primary failure mode in ToMi/ParaToMi.

\paragraph{Step 1: Structural validation.}
We first check whether the generated \protocolname output conforms to the expected schema (e.g., required fields such as timestep/state/observations/actions are present and well-formed). This yields a \textbf{structural score} in $[0,1]$, where higher is better.

\paragraph{Step 2: Observation accuracy via an LLM judge.}
Because the key difficulty in ToMi/ParaToMi is perspective tracking, we focus the judge on whether each agent \emph{observes the right things at the right timesteps}. Concretely, we provide the judge with the story, the question, the formatted ground-truth \protocolname representation, and the formatted generated representation, and ask it to rate \textbf{observation accuracy} (score in $[0,1]$) and provide a brief rationale.
The judge is instructed to prioritize whether agents:
(i) observe events only when they are present in the relevant location,
(ii) do \emph{not} observe events that occur when they are absent,
and (iii) correctly track what they witnessed in prior timesteps.
The judge also considers (as secondary criteria) state accuracy and action attribution.
We use GPT-5 as the judge and request a JSON response containing a numeric score and reasoning text.

\paragraph{Step 3: Composite score.}
We combine the two signals into an overall parser-quality score:
\[
\texttt{overall\_score} = 0.3 \times \texttt{structural\_score} + 0.7 \times \texttt{observation\_accuracy}.
\]
This weighting emphasizes observation fidelity, which most directly affects downstream theory-of-mind reasoning.
We chose the 0.3/0.7 weighting because schema validity is necessary but does not ensure that the represented observations are correct; observation errors directly change who knows what in these tasks. In a manual validation of 24 generated parses from the 100-scenario set, human and judge ratings agreed on 70\% of pairwise comparisons and had Kendall's $\tau=0.40$. Although this is only moderate agreement, both produced the same ranking of parser models, which is the comparison for which we use the judge.

\subsection{Experiment Details}
\label{app:experiment_details}

For tasks that operate under different assumptions about agent perception and knowledge (e.g., in ToMi tasks, agents are assumed to perceive all events occurring within their physical space), we provide task-specific instructions and one exemplar to guide the encoding process.

\subsubsection{Prompt for All Third-person Static Tasks}
Here's the prompt for parsing free-form narratives into \protocolname representations. 

\begin{lstlisting}
Please analyze the following narrative/context.

#### Context: {context}

#### Task specific instructions: {task_specific_instructions}

Example analysis: {example_analysis}

Previous attempt had these issues. 
Please fix them based on the previous attempt and feedback below:
{feedback}

Follow these format instructions:
{format_instructions}
\end{lstlisting}

Here are the task specific instructions for each benchmark:

\paragraph{ToMi}: \texttt{You are dissecting the TOMI scenarios. The assumptions are that the characters can perceive every scene in their location but not scenes occurring elsewhere. If the agent leaves the location, they cannot perceive the scene in that location anymore. In the agent's observation, remember to include the objects' locations if the agents are in the same location as the object.}

\paragraph{HiToM}: \texttt{You are dissecting the HITOM scenarios. You should assume the following: (1) An agent witnesses everything and every movements before exiting a location. (2) An agent A can infer another agent B's mental state only if A and B have been in the same location, or have private or public interactions. (3) Note that every agent tend to lie. What a character tells others doesn't affect his actual belief. (4) Agents in private communications know that others won't hear them, but they know that anyone can hear any public claims. In the agent's observation, remember to include the objects' locations if the agents are in the same location as the object.}

\paragraph{FANToM}: \texttt{You are analyzing a social conversation and need to answer a question about it. When the agents leave the conversation, they cannot perceive the conversation anymore untill they join the conversation again. For convenience, you can use <same\_as\_last\_action /> in the state field to indicate that the state is the same as the last action.}

\paragraph{MMToM-QA}: \texttt{You are dissecting the MMToM scenarios. The assumptions are that agents can perceive objects and events only in their current location. When an agent moves to a new location, they can no longer perceive what happens in previous locations. Importantly, agents should not have knowledge about the contents of containers (like fridges, cabinets, etc.) until they directly observe inside them, unless explicitly stated in their prior knowledge. In mental states, clearly represent the agent's goals, beliefs about object locations, and how these beliefs are updated through observations. In the agent's observation, include objects' locations when the agent is in the same location as the objects, but only after the agent has actually observed them.}

\paragraph{ConfAIde}: \texttt{For convenience, you can use <same\_as\_last\_action /> in the state field to indicate that the state is the same as the last action.}

Here's the prompt for question answering:

\begin{lstlisting}
## Context
{context}
## Extra Info
(to help you better understand the meeting)
{extra_info}
## Task
{question}
\end{lstlisting}

We place \protocolname representations in the extra information entry.

\subsubsection{Model Configurations}
For all experiments, we used the following models:
\begin{itemize}
    \item GPT-4o: \texttt{gpt-4o-2024-08-06}
    \item GPT-4.1: \texttt{gpt-4.1-2025-04-14}
    \item \texttt{o1}: \texttt{o1-2024-12-17}
    \item \texttt{o1}-mini: \texttt{o1-mini-2024-09-12}
    \item \texttt{o3}: \texttt{o3-2025-04-16}
    \item \texttt{o3}-mini: \texttt{o3-mini-2025-01-31}
    \item DeepSeek-R1: \path{together_ai/deepseek-ai/DeepSeek-R1}
    \item Llama-4-Maverick: \texttt{together\_ai/meta-llama/}\allowbreak\texttt{Llama-4-Maverick-17B-128E-}\allowbreak\texttt{Instruct-FP8}
    \item Llama-4-Scout: \path{together_ai/meta-llama/Llama-4-Scout-17B-16E-Instruct}
\end{itemize}

For the experiments in Section~\ref{sec:social_understanding}, we used temperature $0.0$ for all the non-reasoning models (reasoning models do not require temperature).

\subsubsection{Prompt for \texttt{Foresee and Act} Method}
\label{app:forsee_and_act}
Here's the prompt for refining the action:

\begin{lstlisting}[breaklines=true,breakatwhitespace=true]
You are {agent}.
Here is the interaction history between you and the other agent so far:
{history}

Here is your intended action:
{intended_action}

Here is the predicted mental states after you take the intended action 
(you should use them to generate better actions for achieving your goal):
{socialized_context_info}

Please generate a refined action 
so that you can achieve your (i.e., {agent}'s) goal better.

Please only generate a JSON string 
including the action type and the argument.
Your action should follow the given format:
{format_instructions}
\end{lstlisting}

\subsection{Error Analysis Example}
\label{app:error_analysis_example}

\begin{figure}[t]
\centering
\fbox{\begin{minipage}{0.95\columnwidth}
\small
\textbf{Illustrative Example of Social Context Parsing Failure}

\textit{Scenario:} Evelyn places persimmon in basket $\to$ 
Amelia visits garden $\to$ Amelia leaves $\to$ 
Evelyn moves persimmon to bucket

\textit{Question:} Where does Amelia think Evelyn searches for the persimmon?

\textbf{Incorrect parsing:}
\footnotesize
\vspace{0.1em}
\texttt{\{}\\
\texttt{~~"state": "Evelyn moved persimmon to bucket.}\\
\texttt{~~~~~~~~~~~~Amelia is away.",}\\
\texttt{~~"observations": [}\\
\texttt{~~~~"Evelyn: <same\_as\_state />",}\\
\texttt{~~~~"Amelia: <same\_as\_state />" $\leftarrow$ \textbf{WRONG}:}\\
\texttt{~~~~Amelia can't observe events after leaving}\\
\texttt{~~]}\\
\texttt{\}}
\vspace{0.1em}

\small
\textit{Result:} Incorrect parsing leads to wrong prediction 
(\textit{bucket}) instead of correct answer (\textit{basket}).
\end{minipage}}
\caption{Illustrative example of social context parsing failure from error analysis.}
\label{fig:error_analysis}
\end{figure}

\subsection{ConfAIde Benchmark Results}
\label{app:confaide_results}

\paragraph{ConfAIde} \citep{mireshghallah2024can} evaluates inference-time privacy in LLMs. We focus on tier 4 meeting summary tasks, where models must include key details while avoiding disclosure of private information to outsiders. This requires reasoning about each person's knowledge and what information should be shared with different stakeholders. We report the percentage of summaries that satisfy this criterion, using only the meeting summary portion (200 examples) for social reasoning evaluation.

Table~\ref{tab:confaide_results} compares CoT with \protocolname+CoT on ConfAIde. \protocolname+CoT improves four of the seven tested models, with the largest gain for GPT-4o (from 0.575 to 0.740); the mixed results show that the benefit does not hold for every model.

\begin{table}[ht]
    \centering
    \begin{tabular}{lcc}
        \toprule
        \textbf{Model} & \textbf{CoT} & \textbf{\protocolname+CoT} \\
        \midrule
        \texttt{GPT-4o} & 0.575 & 0.740 \\
        \texttt{GPT-4.1} & 0.995 & 0.985 \\
        \texttt{o1} & 0.980 & 0.975 \\
        \texttt{o3} & 0.910 & 0.955 \\
        \texttt{o3-mini} & 0.765 & 0.770 \\
        \texttt{R1} & 0.835 & 0.785 \\
        \texttt{Llama 4} & 0.810 & 0.820 \\
        \midrule
        \textbf{Average} & 0.839 & 0.861 \\
        \bottomrule
    \end{tabular}
    \caption{ConfAIde performance with CoT versus \protocolname+CoT.}
    \label{tab:confaide_results}
\end{table}

\subsection{Additional Experimental Results}
\label{app:add_results}

Table~\ref{tab:social_reasoning_o1} compares CoT with \protocolname+CoT when the representations are generated by the \texttt{o1}-based \protocolname-Parser.

\begin{table}[ht]
    \centering
    \begin{tabular}{lccccc}
        \toprule
        \textbf{Model} & \textbf{ParaToMi} & \textbf{HiToM} & \textbf{MMToM-QA} & \textbf{FANToM} & \textbf{Confaide} \\
        \midrule
        GPT-4o & 0.818 & 0.660 & 0.652 & 0.396 & 0.575 \\
        GPT-4o \small{\textit{\protocolname+CoT}} & 0.885 & 0.750 & 0.692 & 0.472 & 0.625 \\
        \midrule
        GPT-4.1 & 0.802 & 0.830 & 0.586 & 0.528 & 0.995 \\
        GPT-4.1 \small{\textit{\protocolname+CoT}} & 0.892 & 0.760 & 0.583 & 0.623 & 1.000 \\
        \midrule
        \texttt{o1} & 0.835 & 0.870 & 0.725 & 0.415 & 0.980 \\
        \texttt{o1} \small{\textit{\protocolname+CoT}} & 0.933 & 0.880 & 0.761 & 0.528 & 0.990 \\
        \midrule
        \texttt{o3} & 0.955 & 0.930 & 0.715 & 0.547 & 0.910 \\
        \texttt{o3} \small{\textit{\protocolname+CoT}} & 0.947 & 0.890 & 0.722 & 0.698 & 0.975 \\
        \midrule
        \texttt{o3}-mini & 0.817 & 0.810 & 0.493 & 0.057 & 0.780 \\
        \texttt{o3}-mini \small{\textit{\protocolname+CoT}} & 0.863 & 0.790 & 0.460 & 0.151 & 0.750 \\
        \midrule
        R1 & 0.893 & 0.420 & 0.374 & 0.491 & 0.835 \\
        R1 \small{\textit{\protocolname+CoT}} & 0.932 & 0.520 & 0.412 & 0.585 & 0.840 \\
        \midrule
        Llama 4 & 0.740 & 0.720 & 0.443 & 0.264 & 0.810 \\
        Llama 4 \small{\textit{\protocolname+CoT}} & 0.745 & 0.680 & 0.453 & 0.321 & 0.785 \\
        \midrule
        AVG & 0.837 & 0.749 & 0.570 & 0.385 & 0.841 \\
        AVG \small{\textit{\protocolname+CoT}} & 0.885 & 0.753 & 0.583 & 0.483 & 0.852 \\
        \bottomrule
    \end{tabular}
    \caption{CoT versus \protocolname+CoT with the \texttt{o1}-powered \protocolname-Parser.}
    \label{tab:social_reasoning_o1}
\end{table}

\subsubsection{Complete Seven-Model Results with the \texttt{o3} Parser}
Table~\ref{tab:social_reasoning_full} reports the complete grid used for the across-model statistical summary in \S\ref{sec:omniscient_social_reasoning}.
\begin{table}[ht!]
    \centering
    \caption{Performance comparison of models with CoT and \protocolname+CoT representations generated by the \texttt{o3}-powered \protocolname-Parser. Bolded average values indicate the higher method average for each task.}
    
    \adjustbox{max width=\textwidth}{%
    \begin{tabular}{lccccc}
        \toprule
        \textbf{Model} & \textbf{ToMi} & \textbf{ParaToMi} & \textbf{HiToM} & \textbf{FANToM} & \textbf{MMToM-QA} \\
        \midrule
        \texttt{R1} & 0.945 & 0.893 & 0.420 & 0.491 & 0.374 \\
        \texttt{R1} \small{\textit{\protocolname+CoT}} & 0.980 & 0.950 & 0.510 & 0.547 & 0.437 \\
        \midrule
        \texttt{Llama 4}  & 0.655 & 0.740 & 0.720 & 0.264 & 0.443 \\
        \texttt{Llama 4} \small{\textit{\protocolname+CoT}} & 0.662 & 0.763 & 0.700 & 0.415 & 0.450 \\
        \midrule
        \texttt{GPT-4o} & 0.813 & 0.818 & 0.660 & 0.396 & 0.652 \\
        \texttt{GPT-4o} \small{\textit{\protocolname+CoT}} & 0.927 & 0.905 & 0.750 & 0.491 & 0.692 \\
        \midrule
        \texttt{GPT-4.1} & 0.882 & 0.802 & 0.830 & 0.528 & 0.586 \\
        \texttt{GPT-4.1} \small{\textit{\protocolname+CoT}} & 0.978 & 0.913 & 0.810 & 0.585 & 0.583 \\
        \midrule
        \texttt{o1} & 0.952 & 0.835 & 0.870 & 0.415 & 0.725 \\
        \texttt{o1} \small{\textit{\protocolname+CoT}} & 0.985 & 0.932 & 0.880 & 0.623 & 0.785 \\
        \midrule
        
        \texttt{o3-mini} & 0.863 & 0.817 & 0.810 & 0.057 & 0.493 \\
        \texttt{o3-mini} \small{\textit{\protocolname+CoT}} & 0.960 & 0.900 & 0.860 & 0.170 & 0.506 \\
        \midrule
        \texttt{o3} & 0.998 & 0.955 & 0.930 & 0.547 & 0.715 \\
        \texttt{o3} \small{\textit{\protocolname+CoT}} & 0.983 & 0.965 & 0.910 & 0.642 & 0.735 \\

        \midrule
        AVG & 0.873 & 0.837 & 0.749 & 0.385 & 0.570 \\
        AVG \small{\textit{\protocolname+CoT}} & \textbf{0.925} & \textbf{0.904} & \textbf{0.774} & \textbf{0.496} & \textbf{0.598} \\
        \bottomrule
    \end{tabular}
    }
    \label{tab:social_reasoning_full}
\end{table}

\subsubsection{SOTOPIA Additional Results}
\label{app:sotopia_additional}

\begin{figure}[ht]
    \centering
    \includegraphics[width=0.9\textwidth]{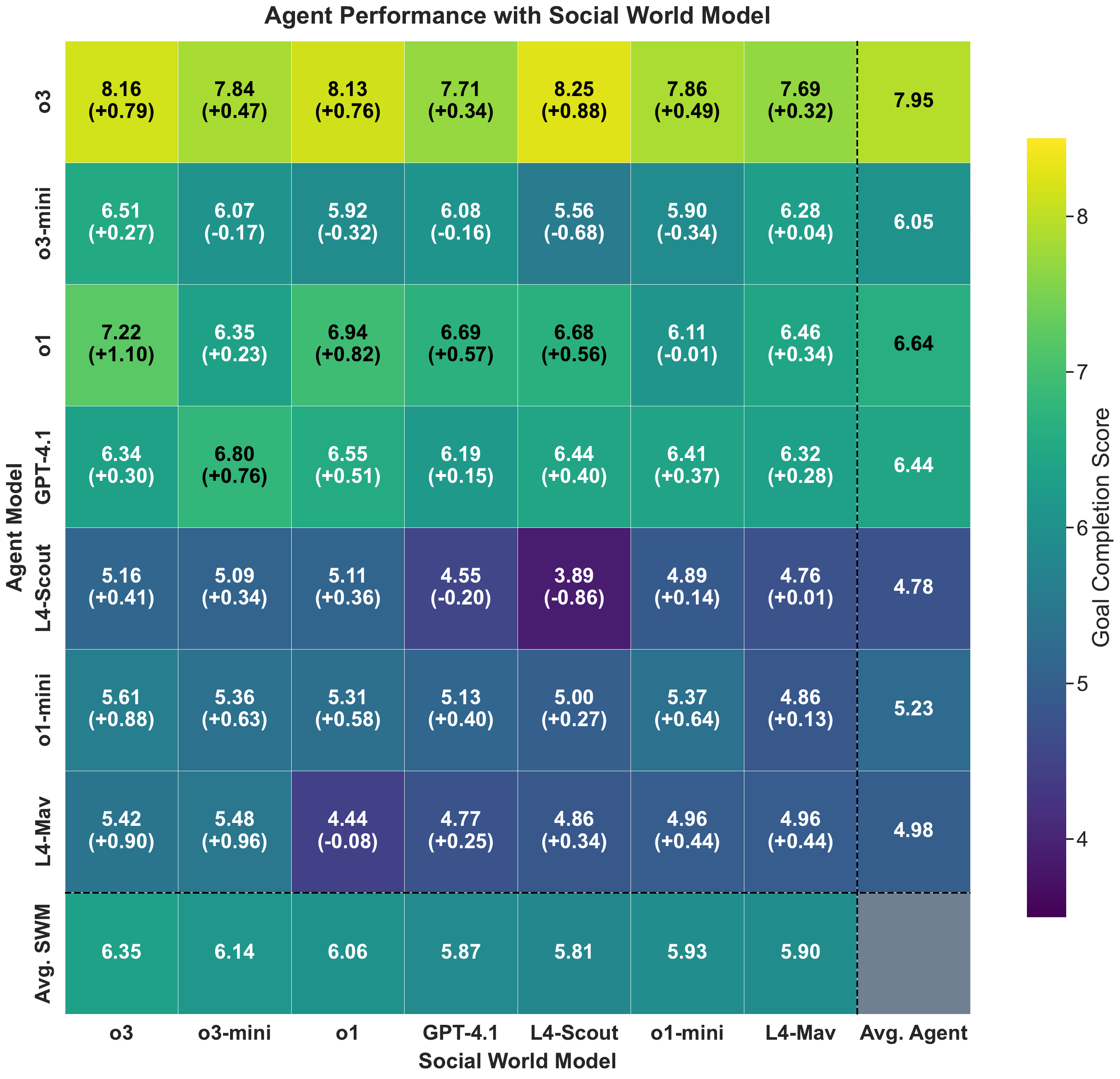}
    \caption{Heatmap showing goal completion scores for different combinations of agent models (rows) and social world models (columns) on SOTOPIA-hard. Values represent goal completion scores (0-10 scale), with darker colors indicating better performance. This visualization reveals which model pairings produce the most effective social reasoning.}
    \label{fig:sotopia_heatmap}
\end{figure}

Figure \ref{fig:sotopia_heatmap} presents a detailed heatmap analysis of model performance across different social world model configurations on the SOTOPIA-hard benchmark. The heatmap visualizes how different combinations of agent models and social world models interact, revealing patterns in which pairings yield the strongest performance improvements.

\begin{figure}[ht]
    \centering
    \includegraphics[width=0.7\textwidth]{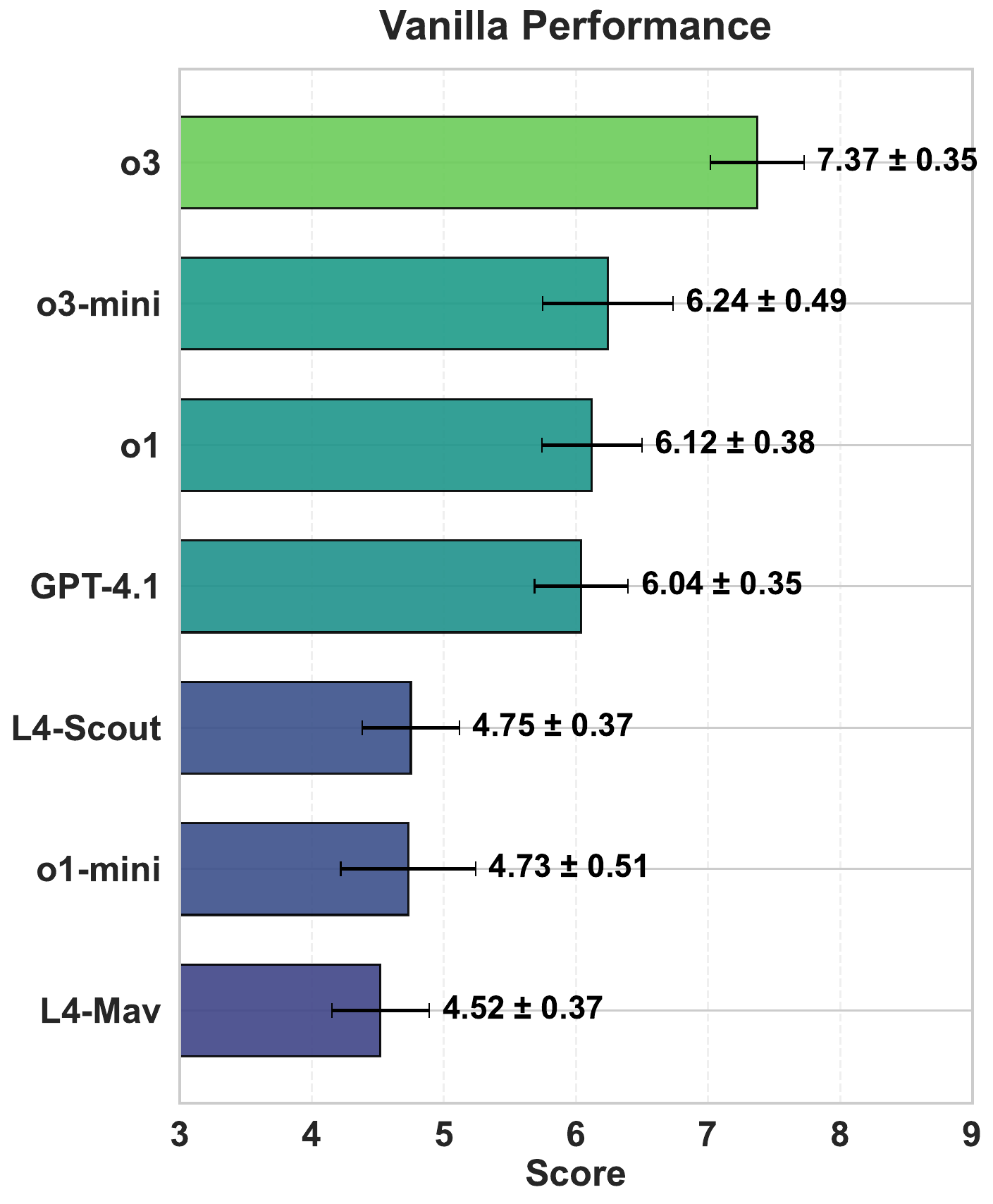}
    \caption{Baseline performance of different models on SOTOPIA-hard without social world model support. Values represent goal completion scores (0-10 scale). These baselines correspond to the right panel in Figure \ref{fig:model_performance_comparison}.}
    \label{fig:sotopia_vanilla}
\end{figure}

Figure \ref{fig:sotopia_vanilla} shows the baseline performance of various models without social world modeling support. This serves as a reference point for understanding the improvements gained through \protocolname-powered social world models.

\subsection{QA Model Scaling Analysis}
\label{app:qwen_scaling}

To examine how QA model size relates to the use of a fixed representation, we evaluate Qwen3 models from 0.6B to 80B parameters on the same 100 randomly sampled ParaToMi instances, using representations generated by \texttt{o3}. \protocolname+CoT improves over CoT at all four tested sizes. The two largest models gain 13--14 points, while the two smaller models gain 1--2 points; this small study suggests a possible size-related pattern but is not sufficient to establish a scaling law.

\begin{table}[ht]
    \centering
    \small
    \caption{ParaToMi accuracy (\%) with CoT versus \protocolname+CoT on Qwen3 models. $\Delta$ (pp) is the absolute performance difference.}
    \label{tab:qwen_effect}
    \begin{tabular}{lccc}
        \toprule
        \textbf{Model} & \textbf{CoT} & \textbf{\protocolname+CoT} & \textbf{$\Delta$ (pp)} \\
        \midrule
        Qwen3-80B   & 77 & 91 & +14 \\
        Qwen3-32B   & 75 & 88 & +13 \\
        Qwen3-7B    & 44 & 45 &  +1 \\
        Qwen3-0.6B  & 48 & 50 &  +2 \\
        \bottomrule
    \end{tabular}
\end{table}

\subsection{Experiments compute resources}
\label{app:compute_resource}

For our experiments, we utilized two main API services for model access:

\begin{itemize}
    \item OpenAI API\footnote{\url{https://platform.openai.com/docs/api-reference}} for accessing GPT-4o, GPT-4.1, \texttt{o1}, \texttt{o1}-mini, \texttt{o3}, and \texttt{o3}-mini models
    \item Together AI API\footnote{\url{https://docs.together.ai/}} for accessing DeepSeek-R1, Llama-4-Maverick, and Llama-4-Scout models
\end{itemize}

All experiments were conducted using these cloud-based APIs, eliminating the need for local GPU resources. The API-based approach allowed us to efficiently scale our experiments while maintaining consistent model access across different benchmarks.

\subsection{Mapping \protocolname to Social World Model Components (Informal)}
\label{app:proof_of_social_world_state}

We provide an informal mapping showing how \protocolname aligns with the components of the social world model defined in Section 3.1. This is \emph{not} a formal completeness proof: any representation is bounded by the information present (or inferable) in the narrative and the parser’s assumptions. The goal here is to clarify what information \protocolname makes explicit and how it can be used to operationalize $\mathcal{S}^t$ in practice:

\begin{enumerate}
    \item \textbf{State Space $\mathcal{S}$}: \protocolname approximates the state space through its structured format containing:
    \begin{itemize}
        \item Environment state $\mathcal{E}^t$ in the \texttt{state} field
        \item Joint observation space $\mathcal{O}^t$ in the \texttt{observations} field
        \item Joint action space $\mathcal{A}^t$ in the \texttt{actions} field
    \end{itemize}
    
    \item \textbf{Observation Space $\mathcal{O}$}: \protocolname captures both external and introspective observations:
    \begin{itemize}
        \item External observations $\mathcal{O}^{\text{ex}}_i$ through direct state descriptions and agent actions
        \item Introspective observations $\mathcal{O}^{\text{in}}_i$ through the \texttt{<mental\_state>} tag
    \end{itemize}
    
    \item \textbf{Memory Function $\Psi$}: \protocolname has no dedicated memory field. Instead, the sequence carries prior observations, actions, and tagged mental states that a model can use as context; this supports history-dependent inference but does not guarantee complete or accurate recall. The available history includes:
    \begin{itemize}
        \item The sequence of simulation steps that can be used to reconstruct $\mathcal{M}_i^t$
        \item Special tags like \texttt{<same\_as\_last\_action>} that maintain temporal consistency
    \end{itemize}
    
    \item \textbf{Transition Function $T$}: The social world model's transition function is approximated through:
    \begin{itemize}
        \item The \texttt{timestep} field that maintains temporal ordering
        \item The sequential nature of simulation steps that captures state transitions
    \end{itemize}
\end{enumerate}

A crucial assumption in our social world model is that agents act independently at each timestep $t$, with no agent having knowledge of others' simultaneous actions. \protocolname supports this assumption through its structured format:
\begin{itemize}
    \item Each agent's actions are recorded separately in the \texttt{actions} field
    \item The \texttt{observations} field captures only what each agent can observe at the current timestep
    \item The sequential nature of simulation steps ensures that agents cannot access future or simultaneous actions
\end{itemize}

Furthermore, \protocolname satisfies key properties of a social world state, though in an approximate manner:

\begin{enumerate}
    \item \textbf{Completeness}: Each simulation step contains all necessary components ($\mathcal{E}^t$, $\mathcal{O}^t$, $\mathcal{A}^t$) to represent a complete social world state at time $t$, though some details may be simplified or omitted.
    
    \item \textbf{Consistency}: The structured format ensures that observations and actions are consistent with the environment state through:
    \begin{itemize}
        \item Special tags that maintain referential integrity
        \item The parser's ability to infer missing elements through reasoning
    \end{itemize}
    
    \item \textbf{Extensibility}: The JSON schema allows for additional metadata and future extensions while maintaining the core social world state representation.
\end{enumerate}

Therefore, \protocolname provides an approximate but practical operationalization of the social world state as defined in our theoretical framework. While it may not capture every nuance of the theoretical model, it offers a structured and computationally tractable way to represent social interactions, and it makes explicit the key fields required to drive downstream reasoning and simulation.

\end{document}